\documentclass[11pt]{article}

\usepackage[final]{acl}

\usepackage{times}
\usepackage{latexsym}
\usepackage[T1]{fontenc}
\usepackage[utf8]{inputenc}
\usepackage{microtype}
\usepackage{inconsolata}

\usepackage{graphicx}
\usepackage{subcaption}
\usepackage{booktabs}
\usepackage{multirow}
\usepackage{wrapfig}
\usepackage{placeins}

\usepackage{amsmath}
\usepackage{amssymb}
\usepackage{mathtools}
\usepackage{amsthm}

\usepackage[ruled,vlined,linesnumbered]{algorithm2e}
\usepackage{algpseudocode}

\usepackage{fancyvrb}
\usepackage{fvextra}
\usepackage{xcolor}
\usepackage{pifont}
\usepackage{verbatim}
\usepackage{fontawesome5}
\usepackage[capitalize,noabbrev]{cleveref}

\title{Demographic Pluralism: Inference-Time Modeling of\\
Pluralistic Human Preference Distributions}
\hypersetup{
  pdftitle={Demographic Pluralism: Inference-Time Modeling of Pluralistic Human Preference Distributions},
  pdfauthor={Meng-Chen Wu, Qipin Chen, Ansh Jain, Tess Wood, Zhe Du, and Si-Chi Chin}
}

\author{
 \textbf{Meng-Chen Wu}\thanks{Equal contribution.},
 \textbf{Qipin Chen}\footnotemark[1],
 \textbf{Ansh Jain},
 \textbf{Tess Wood},
 \textbf{Zhe Du},
 \textbf{Si-Chi Chin}\\
 Amazon Science\\
 \texttt{\{mengchw, qipinche, ajainaj, tesswoo, zddu, sichi\}@amazon.com}
}

\begin{document}
\maketitle

\begin{abstract}
Large language models (LLMs) are increasingly used in culturally sensitive settings, where alignment requires representing diverse preferences within populations.
Yet existing methods model populations at coarse demographic or community levels and overlook within-group variation.
We introduce \textsc{Demographic Pluralism}, an inference-time framework that estimates population-level opinion distributions without opinion-distribution training data or task-specific fine-tuning by generating multiple perspectives within demographically grounded groups.
Across four backbones on GlobalOpinionQA and VITAL, it reduces Jensen--Shannon distance by 8.4\%--26.4\% over Modular Pluralism.
Among weighted, equal-weighted, and inverse-weighted aggregation, equal weighting performs best overall; group-level error also increases with group weight, helping explain weighted aggregation's weaker performance.
\end{abstract}

\section{Introduction}
\label{s:introduction}

Large language models (LLMs) are increasingly used to support decision making in culturally sensitive domains such as policy design, public communication, and health guidance~\citep{singhal2023large, handler2024large, yun2024improving, huang2024collective}.
In these settings, the useful alignment target is often not a single representative answer but the population-level opinion distribution over plausible alternatives.
Consider public health messaging about vaccination: even when the medical evidence is clear, attitudes vary across populations because of differences in institutional trust, religious beliefs, and lived experience.
Observed human opinion distributions can serve as training targets, but they are unavailable for new questions and populations without relevant surveys; we therefore study estimation without using them for training, calibration, or validation.

Estimating population-level opinion distributions without such data presents two challenges.
First, any single group-level representation treats that group as internally homogeneous, even though people sharing the selected demographic profile may differ in education, media exposure, and socioeconomic context.
Second, a method must combine these heterogeneous perspectives into one population-level distribution without treating any group as internally homogeneous.
Many alignment approaches instead optimize toward a dominant or averaged notion of correctness, encouraging models to converge on a single response even when human opinions are widely distributed~\citep{durmus2023towards, tao2024cultural, alkhamissi2024investigating, wu-etal-2025-incorporating}.

Prior work on \emph{pluralistic alignment} represents distributions of perspectives rather than consensus outcomes~\citep{sorensen2024roadmap}.
Modular Pluralism~\citep{feng-etal-2024-modular} represents broad community differences through externally trained community-specific modules, but operates at a coarse level of demographic granularity and does not model variation within each community.
MaxMin-RLHF~\citep{chakraborty2024maxmin} learns policies across conflicting preferences but requires curated, community-specific training data.
\citet{cao2025specializing} fine-tune LLMs using observed opinion distributions as training targets.
\citet{wang2025prompts} introduce Prompts to Proxies (P2P), which keeps the LLM fixed but uses such distributions as targets for training population-specific proxy weights.
Both methods therefore require opinion-distribution training data, although only Cao et al.\ fine-tune the LLM.
Persona-based prompting can elicit diverse viewpoints~\citep{moon-etal-2024-virtual, tseng-etal-2024-two}, but existing methods typically lack both grounding in population structure and a principled mechanism for aggregating perspectives into population-level opinion distributions.

\begin{figure*}[t!]
\centering
\includegraphics[width=\linewidth]{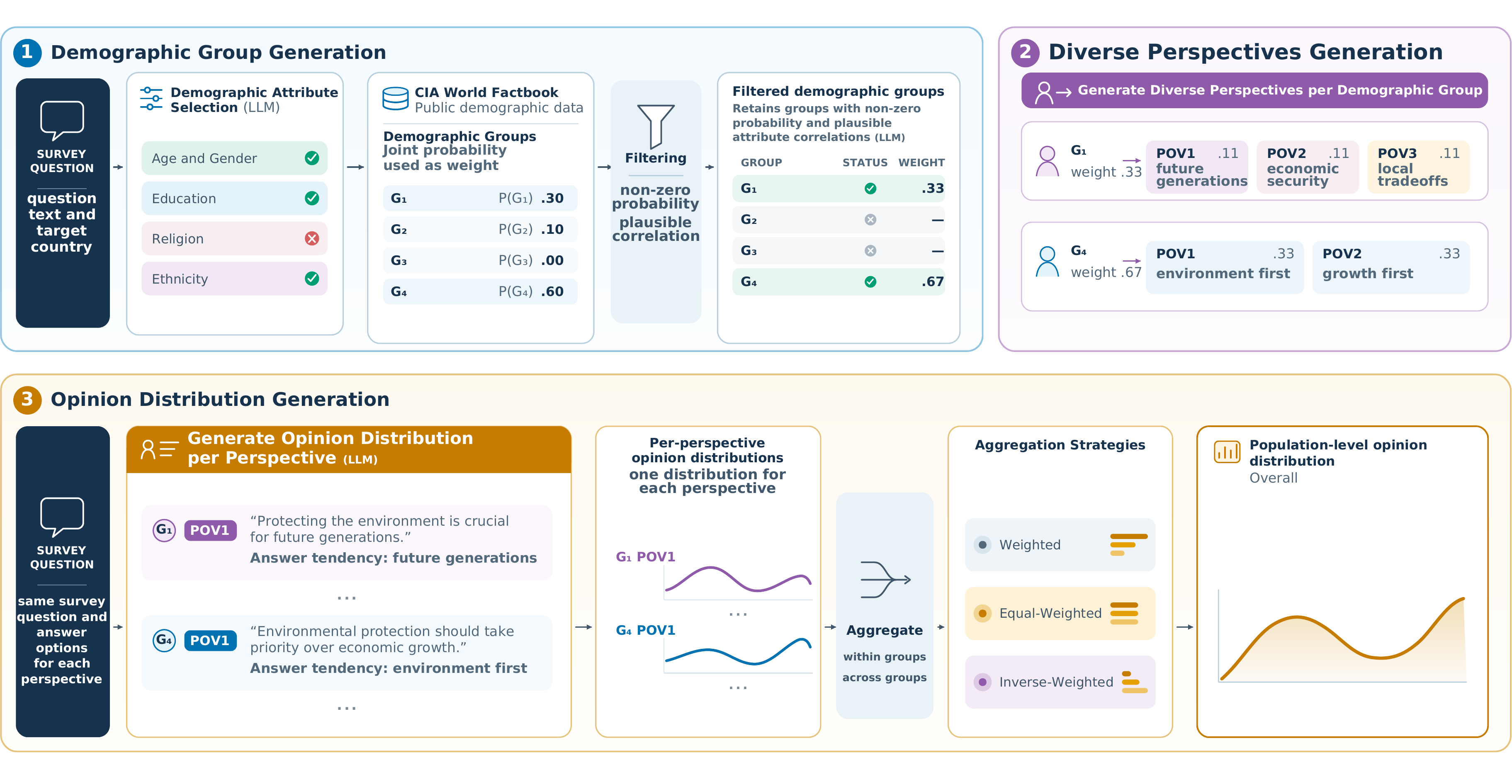}
\caption{
Overview of the \textsc{Demographic Pluralism} framework.
\textbf{(1) Demographic Group Generation} selects question-relevant attributes and constructs plausible demographically supported groups.
\textbf{(2) Diverse Perspectives Generation} elicits multiple viewpoints for each group to capture within-group variation.
\textbf{(3) Opinion Distribution Generation} obtains an answer distribution from each perspective and aggregates the distributions within and across groups.
The framework operates at inference time without observed survey responses or fine-tuning.
}
\label{f:method_flow}
\end{figure*}

We introduce \textsc{Demographic Pluralism}, an inference-time framework for estimating population-level opinion distributions without opinion-distribution training data or method-specific fine-tuning.
Given a survey question and target country, the method selects question-relevant demographic attributes, constructs plausible groups using public demographic statistics, generates multiple perspectives per group, predicts an opinion distribution for each perspective, and aggregates these distributions within and across groups.
These marginals provide population structure only: they characterize population composition but provide no opinion labels or target distributions.
By representing multiple perspectives per group, the framework captures variation both between and within demographic groups.

We evaluate \textsc{Demographic Pluralism} on GlobalOpinionQA~\citep{durmus2023towards} and its health-focused VITAL subset~\citep{shetty-etal-2025-vital} across four open-weight language models using two complementary protocols.
The \emph{full-corpus protocol} evaluates methods that use no observed opinion distributions for training, calibration, or model selection across all 28{,}763 GlobalOpinionQA country--question instances, with VITAL reported separately.
The \emph{question-disjoint protocol} splits GlobalOpinionQA by question into training, validation, and test sets, enabling comparisons with methods that require opinion-distribution training data.
Under the full-corpus protocol, \textsc{Demographic Pluralism} achieves the lowest Jensen--Shannon distance (JSD) for every combination of backbone and evaluation set, reducing JSD by 8.4\%--26.4\% relative to Modular Pluralism.
Under the question-disjoint protocol, it obtains lower JSD than Cao et al.\ on three of four backbones and, in the P2P comparison, lower JSD than P2P on all four backbones.

Our contributions are threefold: (1)~an inference-time framework that uses public demographic data to structure population groups and represents both between- and within-group variation without opinion-distribution training data; (2)~gains over Modular Pluralism under both protocols and lower JSD than methods trained on opinion distributions in most backbone comparisons; and (3)~analyses showing that the gains persist across country resource levels and reconstructed subgroup distributions, while the association between group weight and error helps explain the advantage of equal-weighted aggregation in our experiments.
\section{Methodology}
\label{s:method}

\textsc{Demographic Pluralism} follows the three stages in Figure~\ref{f:method_flow}.
Demographic Group Generation defines a question-relevant population structure before any opinions are elicited (\S\ref{method:demographic_group_generation}).
Diverse Perspectives Generation represents variation within each retained group rather than assigning it one canonical viewpoint (\S\ref{method:within-group-diverse}).
Opinion Distribution Generation then converts each perspective into an answer distribution and separates within-group averaging from cross-group aggregation (\S\ref{method:opinion_distribution}).
This separation lets us compare representational weighting assumptions while holding the generated perspective-level predictions fixed.

\subsection{Demographic Group Generation}
\label{method:demographic_group_generation}

\paragraph{Attributes and Priors.}
We define a fixed inventory containing age and gender, education, religion, and ethnicity.
These attributes cover complementary dimensions of population structure and have public marginal statistics for many countries.
Age and gender are always included because they are standard survey stratification variables~\citep{tourangeau2000psychology,groves2009survey}.
An LLM selects the remaining attributes according to their relevance to question $q$, allowing the demographic representation to vary with the topic rather than imposing the full inventory on every question.\footnote{The selection prompt appears in Appendix~\ref{app:prompt_step1}.}
We obtain country-specific marginal distributions for the selected attributes from public demographic sources.\footnote{Sources and processing are documented in Appendix~\ref{appendix:demographic_data_collection}.}
These statistics provide demographic priors for group construction and aggregation; they neither supervise model outputs nor approximate opinion distributions.

\paragraph{Groups and Joint Weights.}
Let $\mathcal{A}_q=\{A_1,\ldots,A_m\}$ be the attributes selected for $q$, where each $A_k$ takes values in a finite set $\mathcal{V}_k$.
A demographic group $f_i=(a_{i1},\ldots,a_{im})$ assigns one value $a_{ik}\in\mathcal{V}_k$ to every selected attribute.
We enumerate the Cartesian product of these value sets and assign each candidate group the unnormalized demographic weight
\begin{equation}
\bar w_i=\prod_{k=1}^{m}P(A_k=a_{ik}).
\end{equation}
Because only marginals are available in a common form across countries, this product assumes independence among selected attributes.
It supplies a consistent approximation for aggregation, but it does not recover exact intersectional prevalence.

\paragraph{Filtering and Renormalization.}
Enumeration can produce groups that are impossible under the demographic marginals or unlikely because the independence approximation ignores correlations among attributes.
We first remove zero-probability groups deterministically, then use an LLM plausibility filter to remove unlikely nonzero combinations.\footnote{The filtering prompt appears in Appendix~\ref{app:prompt_plausibility}.}
Let $\mathcal{F}'_q$ denote the retained set.
We renormalize its weights as
\begin{equation}
w_i=\frac{\bar w_i}{\sum_{f_\ell\in\mathcal{F}'_q}\bar w_\ell}.
\end{equation}
Filtering determines which groups proceed to perspective generation, while the resulting weights affect only cross-group aggregation.
Neither step supplies an opinion label or constrains a group's generated response distribution.
A complete construction example appears in Appendix~\ref{app:group_example}.

\subsection{Diverse Perspectives Generation}
\label{method:within-group-diverse}

For each question $q$, target country, and retained group $f_i$, generation proceeds in two steps.
We first construct a prototypical persona grounded in the group's measured attributes and country context.
The persona serves as a compact context for elicitation, not as a claim that all members of the group share one biography or opinion.
We then generate $m_i$ distinct perspectives conditioned on both this persona and $q$, so that the variation remains relevant to the decision expressed by the survey question.\footnote{Persona and perspective prompts appear in Appendices~\ref{app:prompt_persona_background} and~\ref{app:prompt_step2}.}

The set $\{p_{i1},\ldots,p_{im_i}\}$ represents plausible variation among people with the same measured demographic attributes.
This design differs from assigning one representative perspective to each group: demographic context localizes the space of views, while multiple generations preserve heterogeneity within that context.
Because no observed within-group response frequencies are available, perspectives are equally weighted within a group.
Consequently, generating more perspectives for one group does not by itself give that group greater population-level influence.
Examples and scale statistics appear in Appendices~\ref{app:within_group_example} and~\ref{appendix:group_statistics}.

\subsection{Opinion Distribution Generation}
\label{method:opinion_distribution}

Each perspective $p_{ij}$ is independently paired with the original survey question and its fixed answer options to produce a perspective-level opinion distribution $P_{ij}(o)$.\footnote{The response-generation prompt appears in Appendix~\ref{app:prompt_step3}.}
This keeps the prediction task fixed across perspectives, with demographic and perspective context providing the varying input.

\paragraph{Within-Group Aggregation.}
Because the generated perspectives have no observed frequencies, we average them uniformly within each group: $D_i(o)=m_i^{-1}\sum_{j=1}^{m_i}P_{ij}(o)$.
This gives every group one distribution and prevents groups with more generated perspectives from gaining influence solely because of their perspective count.

\paragraph{Cross-Group Aggregation.}
We consider three aggregation strategies that encode different assumptions about representation.
\textbf{Weighted aggregation} combines group-level distributions according to their demographic-derived population weights.
\textbf{Equal-weighted aggregation} assigns the same weight to every group, regardless of population size.
\textbf{Inverse-weighted aggregation} uses the normalized complement of each demographic-derived population weight, giving lower-weight groups greater relative influence.
Comparing these strategies allows us to examine how cross-group weighting affects the estimated population-level opinion distribution.
Formal definitions are provided in Appendix~\ref{appendix:aggregation}.
\section{Experiments}
\label{s:experiments}

\subsection{Evaluation Datasets and Metrics}

\paragraph{Datasets.} We evaluate \textsc{Demographic Pluralism} on GlobalOpinionQA and its VITAL subset, both of which provide reference distributions over multiple-choice answers.
GlobalOpinionQA~\cite{durmus2023towards} covers sociopolitical and cultural-value questions across countries.
VITAL~\cite{shetty-etal-2025-vital} is a health-focused, disagreement-filtered subset of GlobalOpinionQA containing healthcare and environment questions.
Detailed dataset descriptions appear in Appendix~\ref{appendix:datasets}.

\paragraph{Evaluation Protocols.}
Following Modular Pluralism~\citep{feng-etal-2024-modular}, we use the same processed GlobalOpinionQA evaluation set, comprising 2{,}088 unique questions and 28{,}763 country--question instances across 131 countries.
We use two complementary protocols that differ in data access and evaluation scope.

\textbf{Full-corpus protocol.}
Observed GlobalOpinionQA opinion distributions are used only for evaluation.
This protocol covers all 28{,}763 instances, with VITAL reported separately.

\textbf{Question-disjoint protocol.}
Following the question-level split in prior work~\citep{cao2025specializing}, we randomly partition the 2{,}088 unique questions into training (35\%), validation (8\%), and test (57\%) sets using seed 42.
The splits contain 730, 167, and 1{,}191 questions (10{,}608, 2{,}262, and 15{,}893 country--question instances), respectively.
All country-specific instances of each question remain in the same split, and methods requiring opinion-distribution training data adapt using only the training split.
\textbf{Panel A} covers the full test set across 131 countries, while \textbf{Panel B} covers the 19-country P2P-compatible test set, containing 3{,}669 instances from 860 questions.

\paragraph{Evaluation Metrics.}
Our primary metric is Jensen--Shannon distance (JSD) between predicted and reference response distributions; lower values indicate better alignment.
JSD treats response options as unordered categories and can therefore be applied consistently to every question.
Because some survey questions have ordinal response options, we additionally report normalized Earth Mover's Distance (EMD) over the full benchmark as a sensitivity analysis.
EMD penalizes probability mass moved farther along the option sequence, although this ordering has a meaningful interpretation only for ordinal questions.
We therefore also report a question-aware Hybrid metric that uses EMD for ordinal questions and JSD for nominal questions.
Appendix~\ref{sec:appendix-metric-comparison} defines these metrics and presents the complete results; all three metrics support the same conclusions.

\subsection{Baselines}

We group baselines by whether adaptation uses opinion-distribution training data, a distinction separate from whether the LLM itself is fine-tuned.

\paragraph{Without opinion-distribution training data.}
These three baselines use no GlobalOpinionQA opinion distributions for training, calibration, or validation.
\textbf{Direct Prompting} conditions the model only on national identity~\cite{feng-etal-2024-modular, tao2024cultural}, treating each country as a homogeneous population.
\textbf{Chain-of-Thought (CoT)}~\cite{wei2022chain} asks the model to consider multiple viewpoints, testing whether deliberation alone improves distributional alignment.\footnote{Prompt templates for Direct Prompting and CoT appear in Appendix~\ref{app:prompt_cot}.}
\textbf{Modular Pluralism}~\cite{feng-etal-2024-modular} uses LLMs fine-tuned on external community corpora to produce one distribution per broad community and averages these distributions equally.
It requires community-specific training but neither uses GlobalOpinionQA opinion distributions nor models within-community variation.
All three baselines are evaluated under both protocols.

\paragraph{With opinion-distribution training data.}
Using GlobalOpinionQA training-set opinion distributions, \textbf{\citet{cao2025specializing}} fine-tune each backbone with LoRA~\citep{hu2022lora} for five epochs under a first-token KL objective, whereas \textbf{P2P}~\citep{wang2025prompts} keeps the LLM fixed and learns sparse, country-specific proxy-agent weights through L1-regularized regression.
Because these weights are country-specific, P2P predicts unseen questions only for calibrated countries.

\subsection{Implementation Details}

We use publicly available open-weight LLMs.
Mistral-7B-Instruct~\cite{jiang2023mistral} generates personas and perspectives; Mistral-7B-Instruct, Phi-3.5-Instruct~\cite{Abdin2024Phi3TR}, Qwen2.5-7B-Instruct-1M~\cite{qwen2.5} (abbreviated Qwen2.5-7B-Instruct), and Qwen3-14B~\cite{qwen3technicalreport} generate opinion distributions.
Following \citet{feng-etal-2024-modular}, distributions are estimated from first-token logits with temperature 0.1 and top-$p$ 0.9.
To account for stochasticity in perspective generation, we repeat the full-corpus evaluation three times and report the mean and standard deviation.
For \textsc{Demographic Pluralism}, run-to-run standard deviations are at most 0.002 JSD (Table~\ref{tab:alignment-results}).
Given this consistency, the question-disjoint comparison uses one run per method on the same fixed split and under the same evaluation conditions.
\section{Results and Discussion}
\label{sec:results}

\begin{table*}[t]
\caption{
\textbf{Jensen--Shannon distance (JSD)} across four language models on GlobalOpinionQA and its VITAL subset (mean over 3 independent runs; standard deviation in parentheses).
\textbf{Bold} marks the best method per model--evaluation-set pair, and \underline{underlining} marks the second-best. Lower is better. Improvement is the relative JSD reduction of the best Demographic Pluralism method over Modular Pluralism.
}
\centering
\small
\renewcommand{\arraystretch}{1.1}
\setlength{\tabcolsep}{3pt}
\resizebox{\textwidth}{!}{%
\begin{tabular}{lcccccccc}
\toprule
\multirow{2}{*}{\textbf{Method}} & 
\multicolumn{2}{c}{\textbf{Mistral-7B-Instruct}} & 
\multicolumn{2}{c}{\textbf{Phi-3.5-Instruct}} & 
\multicolumn{2}{c}{\textbf{Qwen2.5-7B-Instruct}} &
\multicolumn{2}{c}{\textbf{Qwen3-14B}} \\
\cmidrule(lr){2-3} \cmidrule(lr){4-5} \cmidrule(lr){6-7} \cmidrule(lr){8-9}
& Global $\downarrow$ & VITAL $\downarrow$ & Global $\downarrow$ & VITAL $\downarrow$ & Global $\downarrow$ & VITAL $\downarrow$ & Global $\downarrow$ & VITAL $\downarrow$ \\
\midrule
\multicolumn{9}{l}{\textit{Baseline}} \\
Direct Prompting & .3224\,{\scriptsize($5\text{e-}5$)} & .3645\,{\scriptsize($1\text{e-}4$)} & .4178\,{\scriptsize($8\text{e-}5$)} & .3925\,{\scriptsize($5\text{e-}4$)} & .3906\,{\scriptsize($2\text{e-}5$)} & .3576\,{\scriptsize($6\text{e-}4$)} & .4076\,{\scriptsize($8\text{e-}6$)} & .3412\,{\scriptsize($6\text{e-}6$)} \\
CoT & .3905\,{\scriptsize($1\text{e-}3$)} & .3862\,{\scriptsize($4\text{e-}3$)} & .4806\,{\scriptsize($1\text{e-}3$)} & .4639\,{\scriptsize($6\text{e-}3$)} & .3893\,{\scriptsize($6\text{e-}4$)} & .3846\,{\scriptsize($3\text{e-}3$)} & .4234\,{\scriptsize($4\text{e-}4$)} & .4025\,{\scriptsize($4\text{e-}3$)} \\
Modular Pluralism & .3292\,{\scriptsize($7\text{e-}4$)} & .3574\,{\scriptsize($8\text{e-}4$)} & .3366\,{\scriptsize($6\text{e-}4$)} & .3226\,{\scriptsize($2\text{e-}3$)} & .3551\,{\scriptsize($5\text{e-}4$)} & .3462\,{\scriptsize($2\text{e-}3$)} & .3369\,{\scriptsize($1\text{e-}4$)} & .3044\,{\scriptsize($9\text{e-}4$)} \\
\midrule
\multicolumn{9}{l}{\textit{\textsc{Demographic Pluralism}}} \\
\textsc{weighted} & .2884\,{\scriptsize($2\text{e-}4$)} & \textbf{.3275}\,{\scriptsize($6\text{e-}4$)} & .2749\,{\scriptsize($4\text{e-}4$)} & .2843\,{\scriptsize($2\text{e-}3$)} & .2750\,{\scriptsize($2\text{e-}4$)} & .2588\,{\scriptsize($1\text{e-}3$)} & .2835\,{\scriptsize($3\text{e-}4$)} & .2461\,{\scriptsize($1\text{e-}3$)} \\
\textsc{equal-weighted} & \textbf{.2869}\,{\scriptsize($1\text{e-}4$)} & \underline{.3276}\,{\scriptsize($5\text{e-}4$)} & \textbf{.2669}\,{\scriptsize($4\text{e-}4$)} & \textbf{.2808}\,{\scriptsize($5\text{e-}4$)} & \textbf{.2681}\,{\scriptsize($2\text{e-}4$)} & \textbf{.2547}\,{\scriptsize($5\text{e-}4$)} & \textbf{.2757}\,{\scriptsize($2\text{e-}4$)} & \textbf{.2386}\,{\scriptsize($4\text{e-}4$)} \\
\textsc{inverse-weighted} & \underline{.2870}\,{\scriptsize($2\text{e-}4$)} & \textbf{.3275}\,{\scriptsize($6\text{e-}4$)} & \underline{.2670}\,{\scriptsize($4\text{e-}4$)} & \underline{.2815}\,{\scriptsize($2\text{e-}3$)} & \underline{.2682}\,{\scriptsize($2\text{e-}4$)} & \underline{.2553}\,{\scriptsize($1\text{e-}3$)} & \underline{.2758}\,{\scriptsize($2\text{e-}4$)} & \underline{.2394}\,{\scriptsize($2\text{e-}3$)} \\
\midrule
\textbf{Improvement} 
& \textbf{12.8\%} & \textbf{8.4\%}
& \textbf{20.7\%} & \textbf{13.0\%}
& \textbf{24.5\%} & \textbf{26.4\%}
& \textbf{18.2\%} & \textbf{21.6\%} \\
\bottomrule
\end{tabular}%
}
\label{tab:alignment-results}
\end{table*}

\paragraph{Full-Corpus Results.}
Table~\ref{tab:alignment-results} compares methods that use no observed opinion distributions for training, calibration, or model selection.
Across four backbones, \textsc{Demographic Pluralism} achieves the lowest JSD on both GlobalOpinionQA and VITAL, reducing JSD by 8.4\%--26.4\% relative to Modular Pluralism.
These gains persist on VITAL's disagreement-rich health and environmental questions, showing that the improvement is not limited to broad sociocultural topics.

\begin{table*}[!t]
\centering
\caption{
\textbf{Question-disjoint comparison by use of opinion-distribution training data.}
Panel A reports the full test set; Panel B reports the P2P-compatible test set.
Values are JSD ($\downarrow$); \textbf{bold} marks the best result for each backbone.
A black check mark and dash indicate that training-set opinion distributions are used and unused, respectively.
Snowflake and fire symbols indicate that the LLM is frozen and fine-tuned, respectively.
P2P trains only ensemble weights, whereas Modular Pluralism fine-tunes on external community data.
We report equal-weighted \textsc{Demographic Pluralism}, the strongest full-corpus aggregation strategy (see Table~\ref{tab:alignment-results}).
}
\label{tab:heldout-response-supervised}
\small
\renewcommand{\arraystretch}{1.08}
\setlength{\tabcolsep}{4pt}
\resizebox{\textwidth}{!}{%
\begin{tabular}{lcccccc}
\toprule
\textbf{Method}
& \textbf{\shortstack{Uses training-set\\opinion distributions?}}
& \textbf{\shortstack{LLM\\fine-tuned?}}
& \textbf{Mistral-7B}
& \textbf{Phi-3.5}
& \textbf{Qwen2.5-7B}
& \textbf{Qwen3-14B} \\
\midrule

\multicolumn{7}{l}{
\textit{Panel A: Full question-disjoint test (15{,}893 instances)}
} \\
Direct Prompting
& \textcolor{black}{\textemdash}
& \textcolor{cyan!70!blue}{\faSnowflake}
& .327 & .417 & .394 & .406 \\

CoT
& \textcolor{black}{\textemdash}
& \textcolor{cyan!70!blue}{\faSnowflake}
& .393 & .480 & .389 & .423 \\

Modular Pluralism
& \textcolor{black}{\textemdash}
& \textcolor{orange!85!red}{\faFire*}
& .305 & .316 & .308 & .315 \\

Cao et al.
& \textcolor{black}{\faCheck}
& \textcolor{orange!85!red}{\faFire*}
& \textbf{.217} & .362 & .539 & .314 \\

\textsc{Demographic Pluralism}
& \textcolor{black}{\textemdash}
& \textcolor{cyan!70!blue}{\faSnowflake}
& .292 & \textbf{.272} & \textbf{.270} & \textbf{.281} \\

\midrule

\multicolumn{7}{l}{
\textit{Panel B: P2P-compatible test (3{,}669 instances)}
} \\
Direct Prompting
& \textcolor{black}{\textemdash}
& \textcolor{cyan!70!blue}{\faSnowflake}
& .329 & .410 & .395 & .418 \\

CoT
& \textcolor{black}{\textemdash}
& \textcolor{cyan!70!blue}{\faSnowflake}
& .402 & .476 & .395 & .425 \\

Modular Pluralism
& \textcolor{black}{\textemdash}
& \textcolor{orange!85!red}{\faFire*}
& .305 & .313 & .305 & .315 \\

Cao et al.
& \textcolor{black}{\faCheck}
& \textcolor{orange!85!red}{\faFire*}
& \textbf{.215} & .376 & .545 & .320 \\

Prompts to Proxies
& \textcolor{black}{\faCheck}
& \textcolor{cyan!70!blue}{\faSnowflake}
& .360 & .320 & .305 & .290 \\

\textsc{Demographic Pluralism}
& \textcolor{black}{\textemdash}
& \textcolor{cyan!70!blue}{\faSnowflake}
& .291 & \textbf{.267} & \textbf{.265} & \textbf{.279} \\

\bottomrule
\end{tabular}%
}
\end{table*}

\paragraph{Question-Disjoint Results.}
Table~\ref{tab:heldout-response-supervised} shows that, without opinion-distribution training data, \textsc{Demographic Pluralism} achieves lower JSD than Modular Pluralism on all four backbones in both panels.
In \textbf{Panel A}, which reports the full question-disjoint test set, it outperforms Cao et al.\ on three backbones; Cao et al.\ leads only on Mistral-7B (0.217 vs.\ 0.292).
In \textbf{Panel B}, which reports the P2P-compatible test set, it again outperforms Cao et al.\ on three backbones and P2P on all four.
Its Panel A scores also vary less across backbones (0.270--0.292) than those of Cao et al.\ (0.217--0.539), indicating less sensitivity to backbone choice in this evaluation.
These comparison patterns also hold under normalized EMD and the question-aware Hybrid metric (Appendix~\ref{sec:appendix-metric-comparison}), showing that the findings do not depend on treating ordered answer choices as nominal categories.

\subsection{Interpreting Differences Across Methods}
\label{sec:method-differences}

The methods differ primarily in how they represent population structure.
Direct Prompting and CoT predict for each country as a whole and therefore do not explicitly model within-country variation.
Modular Pluralism introduces community-specific modules, but each module produces a single distribution, leaving variation within a community implicit.
Cao et al.\ fine-tune on opinion-distribution training data without explicitly structuring predictions around demographic groups, whereas P2P learns country-specific mixtures over a fixed pool of proxy agents.
In contrast, \textsc{Demographic Pluralism} combines question-relevant demographic groups with multiple perspectives within each group.
These representational differences help explain our improvements, although their contributions vary by benchmark (Section~\ref{sec:component-contributions}) and Cao et al.'s advantage with Mistral-7B shows that no approach performs best across every backbone.

\subsection{Component Contributions}
\label{sec:component-contributions}

To isolate which components of \textsc{Demographic Pluralism} drive the gains, we evaluate three ablations (Appendix~\ref{sec:appendix-ablation}): (1)~removing demographic conditioning while retaining multiple viewpoints tests whether unstructured diversity is sufficient; (2)~generating one perspective per retained group tests the contribution of within-group variation; and (3)~replacing question-dependent attribute selection with a fixed schema tests the value of adapting demographic attributes to each question.

Under equal-weighted aggregation, (1)~removing demographic conditioning generally reduces alignment, except for Mistral-7B on VITAL, indicating that demographic grounding usually contributes beyond unstructured viewpoint diversity.
For (2), using one perspective per group is the most harmful ablation across all four GlobalOpinionQA backbones, showing that within-group variation is particularly important on this broad benchmark.
For (3), using a fixed schema is the most harmful ablation across all four VITAL backbones, indicating that question-dependent attribute selection becomes more important for health-focused questions.
The results therefore do not identify one universally dominant component: demographic grounding generally helps, while the relative importance of within-group variation and adaptive attribute selection depends on the benchmark.

\subsection{Effects of Aggregation Strategies}
\label{sec:aggregation-effects}

In every setting in Table~\ref{tab:alignment-results}, at least one rebalanced strategy, equal- or inverse-weighted aggregation, matches or outperforms weighted aggregation, with equal weighting producing the strongest overall results.
To understand this pattern, we analyze prediction error at the group level and find that greater group weight is associated with higher JSD, particularly in lower-resource GlobalOpinionQA countries and across VITAL resource tiers (Appendix~\ref{appendix:aggregation-bias}).
Under these observed error patterns, weighted aggregation assigns greater influence to less accurate group estimates, whereas the rebalanced strategies limit this concentration.
This association helps explain the advantage of equal weighting in our experiments, but it neither establishes why higher-weight groups have greater errors nor implies that equal weighting is universally optimal.
\section{Analysis}
\label{s:analysis}

Beyond aggregate scores, we examine majority-default collapse, country-resource variation, subgroup fidelity, model scale, and inference cost. For consistency, we use \texttt{Mistral-7B-Instruct} with equal-weighted aggregation unless noted.

\subsection{Mitigating Majority-Default Collapse}
\label{sec:majority-default}
We define \emph{majority-default collapse} as concentrating probability on one response option even when the reference distribution assigns substantial mass to alternatives.
For each question--country pair, we define \emph{gold dominance} as the largest reference probability.
Pairs with dominance at or below 60\% are pluralistic, while those above 60\% are dominant.
Under this definition, 56.8\% of GlobalOpinionQA pairs and 41.8\% of VITAL pairs have no clear majority, making distributed opinion the common case rather than an edge condition.

\begin{figure*}[t]
\centering
\includegraphics[width=0.75\textwidth]{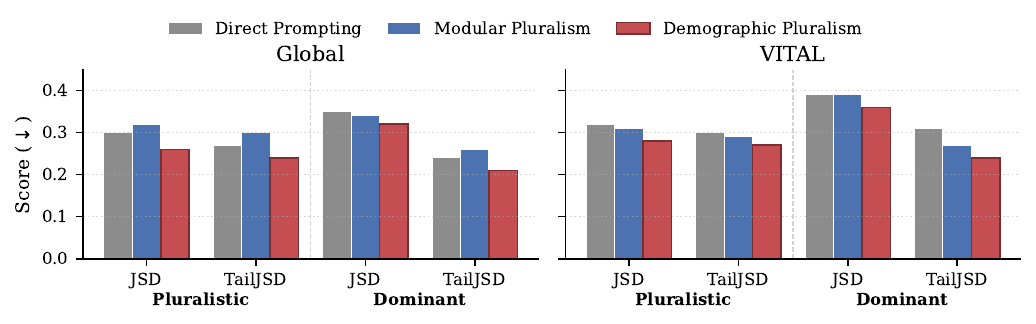}
\caption{
Alignment on pluralistic (gold dominance $\leq 0.60$) and dominant ($>0.60$) cases.
\textsc{Demographic Pluralism} achieves the lowest JSD and TailJSD in both regimes, with the largest gaps on pluralistic cases where baselines collapse onto a single option.
Lower is better; Top-2 mass is reported in Appendix~\ref{sec:appendix-threshold}.
}
\label{f:pluralism-collapse}
\end{figure*}

Figure~\ref{f:pluralism-collapse} compares overall JSD and TailJSD over non-dominant options; Top-2 mass is reported in Appendix~\ref{sec:appendix-threshold}.
On pluralistic cases, Direct Prompting and Modular Pluralism concentrate too much mass on one response and underrepresent alternatives supported by the reference distribution.
\textsc{Demographic Pluralism} assigns more mass beyond the dominant option and achieves lower JSD and TailJSD, indicating that the added probability is placed on relevant alternatives rather than spread indiscriminately.
On dominant cases, predictions remain similarly concentrated across methods.
The gains therefore arise primarily from mitigating collapse when disagreement is present, not from injecting uncertainty into questions with strong consensus.
The same pattern holds under alternative dominance thresholds (Appendix~\ref{sec:appendix-threshold}).

\paragraph{Qualitative Analysis.}
A GlobalOpinionQA case study illustrates how the aggregate difference emerges (Appendix~\ref{appendix:qualitative}).
The reference distribution has no clear majority, with a maximum probability of $0.30$.
Direct Prompting and Modular Pluralism nevertheless concentrate on one option, yielding JSD values of $0.169$ and $0.177$.
In contrast, demographic-grounded perspectives emphasize distinct considerations related to age, gender, and religion, causing their perspective-level distributions to support different plausible responses.
Averaging these distributions produces a population estimate with JSD $0.019$.
This example shows the intended mechanism: demographic context structures the diversity, while multiple perspectives allow disagreement within the same group to affect the final distribution.

\subsection{Impact of Country Resource Levels}
\label{sec:resource_impact}

\begin{figure}[t]
\centering
\includegraphics[width=\columnwidth]{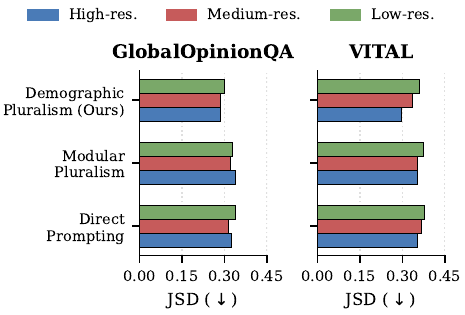}
\caption{
JSD ($\downarrow$) by resource tier on GlobalOpinionQA (left) and VITAL (right). \textsc{Demographic Pluralism} is lowest in every tier.
}
\label{f:resource_level}
\end{figure}

We examine alignment across countries grouped into high-, medium-, and low-resource tiers based on World Bank income categories, UNDP HDI ratings, and the IMF Advanced Economies List~\cite{worldbank_income, undp_hdi, imf_monetary}. Because Direct Prompting and Modular Pluralism are the most competitive baselines in the full-corpus protocol, we compare against these two methods in this analysis. Baseline JSD differs significantly across tiers in most method--benchmark settings (Table~\ref{tab:anova-resource}), although the direction is not uniform. As shown in Figure~\ref{f:resource_level}, \textsc{Demographic Pluralism} achieves the lowest JSD in every tier on both benchmarks, reducing JSD by 8.8\%--15.6\% on GlobalOpinionQA and 3.4\%--16.1\% on VITAL relative to Modular Pluralism, with gains persisting in low-resource countries. Detailed statistics and ANOVA results are provided in Appendix~\ref{appendix:stats_country_resource}.

\subsection{Subgroup-Level Validation}
\label{sec:subgroup_eval}

Aggregate agreement alone cannot determine whether a method captures demographic variation or merely produces errors that cancel at the country level.
We therefore reconstruct subgroup reference distributions from respondent microdata in the Pew Global Attitudes Survey (2002--2022)~\citep{pew_global_morality} and World Values Survey Wave~7, version 6.0.0~\citep{haerpfer2022wvs}.
For each available question--group pair, we aggregate responses from respondents matching the modeled age, gender, education, religion, and ethnicity attributes.
This procedure yields references for 724{,}395 of 917{,}876 groups (78.9\%).
The uncovered groups belong to 17 countries without public microdata, so this analysis validates the covered subset rather than all modeled populations.

Table~\ref{tab:subgroup-validation} compares the full method with an ablation that retains the same demographic conditioning but generates only one perspective per group, using Qwen2.5-7B-Instruct with equal-weighted aggregation on the microdata-covered subset.
The comparison therefore isolates the contribution of within-group multiplicity after group construction is held fixed.
Multiple perspectives reduce subgroup JSD by 29.9\%, close to the 32.9\% reduction at the country level.
They also strengthen the correspondence between country- and subgroup-level performance: Pearson correlation rises from $0.695$ to $0.799$, while Spearman correlation rises from $0.699$ to $0.779$.
Together, these measurements show that the aggregate improvement extends to observed demographic intersections rather than appearing only after country-level averaging.
The evidence supports subgroup opinion-distribution fidelity, but it does not establish that every generated group is plausible, that product-of-marginals weights recover exact joint prevalence, or that generated personas are free of stereotypes.


\begin{table}[t]
\caption{
Country- and subgroup-level alignment on the microdata-covered GlobalOpinionQA subset using Qwen2.5-7B-Instruct and equal-weighted aggregation. The full method improves over the ablation at both levels and yields a stronger country--subgroup correlation. Lower JSD is better.
}
\centering
\small
\renewcommand{\arraystretch}{1.15}
\setlength{\tabcolsep}{3pt}
\resizebox{\columnwidth}{!}{%
\begin{tabular}{lcccc}
\toprule
\textbf{Method}
& \textbf{Country} & \textbf{Subgroup}
& \textbf{Pearson} & \textbf{Spearman} \\
& \textbf{JSD $\downarrow$} & \textbf{JSD $\downarrow$} & & \\
\midrule
\textsc{Demographic Pluralism}
& \textbf{0.2106} & \textbf{0.2671} & \textbf{0.799} & \textbf{0.779} \\

\textit{w/o Multiple Perspectives}
& 0.3139 & 0.3811 & 0.695 & 0.699 \\

\bottomrule
\end{tabular}%
}
\label{tab:subgroup-validation}
\end{table}

\subsection{Robustness to Intermediate Model Scale}
\label{sec:scaling}

We next ask whether the result depends on the particular intermediate model used to construct personas and perspectives.
Holding the opinion model fixed at Qwen2.5-7B-Instruct, we replace Mistral-7B-Instruct with Qwen3-32B~\citep{qwen3technicalreport}.
The larger generator does not improve alignment: JSD rises from $0.2681$ to $0.2788$ on GlobalOpinionQA and from $0.2547$ to $0.3095$ on VITAL, although both configurations remain substantially better than CoT (Appendix~\ref{sec:appendix-scaling}).
At the same time, embeddings over 665{,}264 matched question--group pairs show that the two generators produce substantially different perspectives: cross-model cosine similarity is approximately $0.50$, and Qwen3-32B produces greater within-model diversity (Table~\ref{tab:artifact-similarity}).
The absence of gains from the larger generator, despite these different intermediate realizations, supports the structural role of demographic decomposition and within-group aggregation rather than dependence on richer wording.
This experiment covers two generator scales and does not establish invariance to arbitrary models or prompting choices.

\subsection{When Additional Attributes Matter}
\label{sec:attribute_domain_dependence}

The whole-benchmark ablation shows that adaptive selection matters differently by benchmark: a fixed schema is the most harmful ablation for every VITAL backbone, but has smaller effects on GlobalOpinionQA, where using one perspective per group is most harmful (Appendix~\ref{sec:appendix-ablation}).
A matched-subset diagnostic with Qwen2.5-7B-Instruct sharpens this difference: relative to age and gender alone, adding religion and ethnicity improves JSD by $0.199$ on VITAL but $0.023$ on GlobalOpinionQA; adding education changes these gains to $0.200$ and $0.024$.
VITAL also shows greater cross-country schema variation (entropy $0.989$ vs.~$0.858$; $3.53$ vs.~$3.03$ combinations per question).
Although not directly comparable to the whole-benchmark adaptive ablation, this diagnostic supports question-adaptive selection for health-focused questions without establishing causal effects of any attribute (Appendix~\ref{sec:appendix-attribute-domain}).

\subsection{Runtime Analysis}
\label{sec:runtime}

Generating several perspectives and opinion distributions requires more calls than Direct Prompting.
On GlobalOpinionQA, the measured full-pipeline runtime is $61{,}884\,\mathrm{s}$, close to Modular Pluralism ($59{,}856\,\mathrm{s}$) and below CoT ($73{,}562\,\mathrm{s}$).
On the smaller VITAL subset, the method takes $6{,}834\,\mathrm{s}$, compared with $2{,}185\,\mathrm{s}$ for Modular Pluralism and $4{,}842\,\mathrm{s}$ for CoT, reflecting the fixed cost of constructing groups and perspectives (Appendix~\ref{appendix:runtime}).

Because demographic weights are known after group construction, the pipeline can rank candidate perspectives by population weight and generate only the top-$K$ perspectives and their opinion distributions.
The post-hoc top-$K$ analysis retains strong quality at $K=5$ and generally stabilizes by $K=20$--$30$ (Appendix~\ref{appendix:minimal_personas}).
These results establish an accuracy--coverage tradeoff, but direct top-$K$ wall-clock savings have not yet been measured.
Finally, because each completed perspective yields an independent distribution, perspectives can be pruned and aggregation weights changed post-hoc without further model inference.

\section{Related Work}
\label{s:related_works}

\paragraph{Cultural Alignment.}
LLM research and evaluation often overrepresent values from Western, Educated, Industrialized, Rich, Democratic (WEIRD) societies~\citep{atari2023which,durmus2023towards}, while generated outputs can converge toward algorithmic monoculture~\citep{wu2024generative,zhang2026community}.
Prior methods use cultural prompting~\citep{alkhamissi2024investigating,tao2024cultural}, soft prompt tuning~\citep{masoud2025soft}, or culture-specific adaptation~\citep{li2024culturellm}, but generally target one culture rather than a population-level opinion distribution across demographic groups.

\paragraph{Pluralistic Alignment.}
Pluralistic alignment accommodates conflicting value systems rather than one averaged preference~\citep{sorensen2024roadmap,chakraborty2024maxmin}.
Approaches include participatory aggregation~\citep{huang2024collective}, social-choice analyses of preference optimization~\citep{golz2025distortion}, and group-specific preference learning~\citep{srewa2025pluralllm}.
Modular Pluralism uses externally trained community modules~\citep{feng-etal-2024-modular} but does not model within-community variation; our setting instead estimates population-level opinion distributions at inference time without opinion-distribution training data.

\paragraph{Persona Modeling.}
Persona methods elicit viewpoints through role-playing, backstories, fine-tuning, or multi-agent simulation~\citep{tseng-etal-2024-two,moon-etal-2024-virtual,jandaghi-etal-2024-faithful,hou2025can}; PERSONA and PersonaGym evaluate persona fidelity~\citep{castricato2024persona,samuel-etal-2025-personagym}.
\textsc{Demographic Pluralism} instead uses personas as intermediate representations, grounds groups in demographic statistics, models multiple within-group perspectives, and aggregates their predictions into a population-level opinion distribution.

\section{Conclusion}
\label{s:conclusion}

We propose \textsc{Demographic Pluralism}, an inference-time framework that estimates population-level opinion distributions without opinion-distribution training data by modeling variation across and within demographic groups.
Across four backbones, it reduces JSD by 8.4\%--26.4\% over Modular Pluralism; under question-disjoint evaluation, it achieves lower JSD than Cao et al.\ on three backbones and P2P on all four within the 19-country P2P-compatible test set.
Ablations identify within-group variation as most consequential on GlobalOpinionQA and adaptive attribute selection on VITAL; per-perspective outputs enable top-$K$ control and reweighting.

\section*{Limitations}
\label{sec:limitations}

Our evaluation covers the complete GlobalOpinionQA corpus and its disagreement-filtered VITAL subset, spanning broad sociocultural topics and focused health and environment questions within a shared source corpus. This design varies question domain while holding the survey source constant; evaluation on independently collected datasets would test how broadly the findings transfer. The task is population-level distribution estimation rather than individual-level prediction. The product-of-marginals approximation does not recover exact intersectional prevalence, and public demographic-data coverage varies across countries. Equal weighting performs best in these experiments but is not a universal fairness rule or a substitute for an application-specific estimand.

Attribute selection, plausibility filtering, persona construction, and perspective generation all depend on LLM priors and may reproduce stereotypes. Subgroup evaluation tests opinion-distribution fidelity only for covered question--group pairs; it neither validates every generated group and joint weight nor eliminates representational harm. Broader datasets, participatory evaluation, and direct runtime measurement of top-$K$ generation remain future work.

\section*{Ethics Statement}\label{sec:ethics}

The framework estimates aggregate opinion distributions from fictitious personas; it is not designed to infer, track, or predict any identifiable person's beliefs. Its outputs should not replace surveys, customer research, or consultation with affected communities, especially in policy or health decisions. Demographic attributes must not be used to target individuals or justify decisions about protected groups.

Demographic conditioning can reproduce stereotypes, and an LLM plausibility filter can exclude valid minority intersections. Subgroup evaluation provides limited evidence about distributional fidelity but does not remove these risks. Deployments should disclose the generated nature of the estimates, audit subgroup harms, validate the aggregation rule for the intended population, minimize sensitive-data use, and include participatory review.

\section*{Acknowledgments}
Pew Research Center bears no responsibility for the analyses or interpretations of the data presented here. The opinions expressed herein, including any implications for policy, are those of the author and not of Pew Research Center.

\bibliography{custom}

\appendix
\appendix

\section{Demographic Data Collection}
\label{appendix:demographic_data_collection}

The \textsc{Demographic Pluralism} framework leverages public demographic statistics to construct demographically grounded contexts for pluralistic opinion modeling. These data are used \emph{only} to define feasible demographic groups and to estimate their relative population weights; they are never used to supervise, approximate, or constrain opinion distributions. This design ensures demographic grounding without imposing normative assumptions about group-level preferences.

We collect country-level demographic data for five core attributes: age, gender, education, religion, and ethnicity. These attributes span age, gender, educational, religious, and ethnic dimensions of population diversity and are broadly available across countries and regions. Age and gender are included as default attributes for all survey questions, while education, religion, and ethnicity are selected in a query-dependent manner based on relevance, as described in Section~\ref{s:method}. Together, these attributes provide a flexible yet structured basis for demographic localization.

Demographic attributes are combined to form candidate demographic groups under an independence assumption. Two filtering steps are then applied: (1) groups with zero empirical probability under the source marginals are removed, and (2) statistically non-zero but demographically implausible combinations (e.g., ethnicity--religion pairings that are rarely observed within a given country) are filtered using the model's prior knowledge. After filtering, joint probabilities are renormalized to sum to one. The resulting set of demographic groups represents feasible and non-negligible subpopulations used for subsequent perspective generation and aggregation.

Below, we describe each demographic attribute and the corresponding data sources used in our experiments.

\begin{itemize}

\item \textbf{Age and gender} --- Joint population distributions over age and gender, capturing demographic variation across life stages. Age bands follow the source statistics: 15--64 years and older adults aged 65 years or above. The first band includes ages 15--17 and should therefore not be interpreted as an adults-only category. Gender proportions are modeled jointly within each age group, allowing demographic groups to reflect age-specific gender composition observed in the population. Data are sourced from the CIA World Factbook\footnote{\url{https://www.cia.gov/the-world-factbook/field/age-structure/}}.

\item \textbf{Educational attainment} --- Educational attainment distributions among adults aged 25--64 years, categorized according to the International Standard Classification of Education framework. The percentages come from Indicator A1 of OECD's \emph{Education at a Glance 2023}~\citep{oecd2023education}. These data enable the creation of personas with diverse educational backgrounds that reflect country-specific patterns:

\begin{itemize}

\item \textbf{Tertiary education}: University-level and specialized higher education, including colleges, technical institutes, community colleges, and research centers.

\item \textbf{Upper secondary or post-secondary non-tertiary education}: Final stage of secondary education preparing for higher education or employment, featuring specialized subjects, or programs bridging between secondary education and either tertiary education or the labor market.

\item \textbf{Below upper secondary education}: Educational attainment below the upper secondary level, including primary education, lower secondary education, and no formal education.

\end{itemize}

\item \textbf{Religious affiliation} --- Country-level distributions of religious affiliation across major faith traditions (Christianity, Islam, Hinduism, Buddhism, Folk religions, Judaism) and secular identifications (unaffiliated), as well as smaller traditions (e.g., Sikhism, various forms of paganism). This data ensures personas reflect the religious diversity that shapes cultural perspectives. Data are sourced from the CIA World Factbook\footnote{\url{https://www.cia.gov/the-world-factbook/field/religions/}} 
 via Wikipedia\footnote{\url{https://en.wikipedia.org/wiki/Religions_by_country}}.

\item \textbf{Ethnicity} --- Distributions of major ethnic groups within each country, capturing cultural and linguistic diversity essential for representative persona creation. This demographic dimension helps model perspectives shaped by ethnic identity and cultural heritage. Data are compiled from Wikipedia\footnote{\url{https://en.wikipedia.org/wiki/List_of_countries_by_ethnic_groups}} based on CIA World Factbook estimates and national census data.

\end{itemize}

\section{Aggregation Strategies}
\label{appendix:aggregation}

Let $\mathcal{F}' = \{f_i\}_{i=1}^{n}$ denote the retained demographic groups with renormalized weights $w_i$, and let $\{p_{i1},\ldots,p_{im_i}\}$ denote the $m_i$ perspectives generated within group $f_i$ (Section~\ref{method:within-group-diverse}). Each perspective $p_{ij}$ produces an opinion distribution $P_{ij}(o)$ over answer options $o$. We consider three strategies for aggregating these distributions into a final opinion distribution $P(o)$:

\begin{itemize}
\item \textbf{Weighted aggregation}:
\begin{equation}
P(o) = \sum_{i=1}^{n} w_i \cdot \left( \frac{1}{m_i} \sum_{j=1}^{m_i} P_{ij}(o) \right)
\end{equation}

\item \textbf{Equal-weighted aggregation}:
\begin{equation}
P(o) = \frac{1}{n} \sum_{i=1}^{n} \left( \frac{1}{m_i} \sum_{j=1}^{m_i} P_{ij}(o) \right)
\end{equation}

\item \textbf{Inverse-weighted aggregation}:
\begin{equation}
\begin{aligned}
P(o) &= \sum_{i=1}^{n}
\frac{1-w_i}{\sum_{\ell=1}^{n}(1-w_\ell)} \\
&\quad \cdot \left(\frac{1}{m_i}\sum_{j=1}^{m_i}P_{ij}(o)\right).
\end{aligned}
\end{equation}
\end{itemize}

Weighted aggregation mirrors real-world population proportions, equal-weighted aggregation balances group contributions regardless of prevalence, and inverse-weighted aggregation reallocates mass toward smaller groups using complement-based weights.

\section{Additional Method Details}
\label{appendix:more_method_details}
\subsection{Example of Demographic Group Construction}
\label{app:group_example}

We provide a complete example of the final output produced by the demographic group generation procedure described in Section~\ref{method:demographic_group_generation}. For a survey question asked in Belgium regarding Germany's influence in the European Union, the selected demographic attributes are age and gender (default), education, and religion. After enumerating attribute combinations and filtering zero-probability or demographically implausible combinations, the method produces the following retained joint weights before the final normalization in Equation~(2).

\noindent\textbf{Example of demographic group generation for a survey question about Germany's influence in the EU:}
\begin{Verbatim}[fontsize=\scriptsize,breaklines=true,breakanywhere=true]
{
    "country": "Belgium",
    "question": "Question: When it comes to Germany's decision-making in the European Union, do you think Germany has too much influence, has too little influence or has about the right amount of influence?\nOptions: A. Has too much influence B. Has too little influence C. Has about the right amount of influence D. DK/Refused\nAnswer:",
    "attributes": [
        "age_and_gender",
        "education",
        "religion"
    ],
    "joint_distribution": {
        "1": 0.11218697503789998,
        "3": 0.050636922585199994,
        "12": 0.08988269248384999,
        "13": 0.00825381923198,
        "14": 0.0405696199538,
        "23": 0.04304408247825,
        "24": 0.003952686173100001,
        "25": 0.019428457461,
        "34": 0.11004027800572,
        "36": 0.049667985403359996,
        "45": 0.08816278775217999,
        "46": 0.008095882455064,
        "47": 0.03979332054184,
        "56": 0.0422204342421,
        "57": 0.0038770515490800005,
        "58": 0.0190566940548,
        "67": 0.031847573778779996,
        "69": 0.014374780382639998,
        "78": 0.025515846909569997,
        "80": 0.01151688031716,
        "89": 0.01221932931165,
        "91": 0.0055153393002000005,
        "100": 0.039993259777599995,
        "102": 0.018051432428799998,
        "111": 0.0320420607544,
        "113": 0.014462564387199998,
        "118": 0.00014961273504,
        "122": 0.015344679468000002,
        "124": 0.006926003184
    },
    "formatted_combinations": {
        "1": "Age_And_Gender: 15-64 years (male)\nEducation: Tertiary\nReligion: Christian",
        "3": "Age_And_Gender: 15-64 years (male)\nEducation: Tertiary\nReligion: Irreligion",
        "12": "Age_And_Gender: 15-64 years (male)\nEducation:  Upper secondary or post-secondary non-tertiary\nReligion: Christian",
        "13": "Age_And_Gender: 15-64 years (male)\nEducation:  Upper secondary or post-secondary non-tertiary\nReligion: Muslim",
        "14": "Age_And_Gender: 15-64 years (male)\nEducation:  Upper secondary or post-secondary non-tertiary\nReligion: Irreligion",
        "23": "Age_And_Gender: 15-64 years (male)\nEducation:  Below upper secondary\nReligion: Christian",
        "24": "Age_And_Gender: 15-64 years (male)\nEducation:  Below upper secondary\nReligion: Muslim",
        "25": "Age_And_Gender: 15-64 years (male)\nEducation:  Below upper secondary\nReligion: Irreligion",
        "34": "Age_And_Gender: 15-64 years (female)\nEducation: Tertiary\nReligion: Christian",
        "36": "Age_And_Gender: 15-64 years (female)\nEducation: Tertiary\nReligion: Irreligion",
        "45": "Age_And_Gender: 15-64 years (female)\nEducation:  Upper secondary or post-secondary non-tertiary\nReligion: Christian",
        "46": "Age_And_Gender: 15-64 years (female)\nEducation:  Upper secondary or post-secondary non-tertiary\nReligion: Muslim",
        "47": "Age_And_Gender: 15-64 years (female)\nEducation:  Upper secondary or post-secondary non-tertiary\nReligion: Irreligion",
        "56": "Age_And_Gender: 15-64 years (female)\nEducation:  Below upper secondary\nReligion: Christian",
        "57": "Age_And_Gender: 15-64 years (female)\nEducation:  Below upper secondary\nReligion: Muslim",
        "58": "Age_And_Gender: 15-64 years (female)\nEducation:  Below upper secondary\nReligion: Irreligion",
        "67": "Age_And_Gender: 65+ years (male)\nEducation: Tertiary\nReligion: Christian",
        "69": "Age_And_Gender: 65+ years (male)\nEducation: Tertiary\nReligion: Irreligion",
        "78": "Age_And_Gender: 65+ years (male)\nEducation:  Upper secondary or post-secondary non-tertiary\nReligion: Christian",
        "80": "Age_And_Gender: 65+ years (male)\nEducation:  Upper secondary or post-secondary non-tertiary\nReligion: Irreligion",
        "89": "Age_And_Gender: 65+ years (male)\nEducation:  Below upper secondary\nReligion: Christian",
        "91": "Age_And_Gender: 65+ years (male)\nEducation:  Below upper secondary\nReligion: Irreligion",
        "100": "Age_And_Gender: 65+ years (female)\nEducation: Tertiary\nReligion: Christian",
        "102": "Age_And_Gender: 65+ years (female)\nEducation: Tertiary\nReligion: Irreligion",
        "111": "Age_And_Gender: 65+ years (female)\nEducation:  Upper secondary or post-secondary non-tertiary\nReligion: Christian",
        "113": "Age_And_Gender: 65+ years (female)\nEducation:  Upper secondary or post-secondary non-tertiary\nReligion: Irreligion",
        "118": "Age_And_Gender: 65+ years (female)\nEducation:  Upper secondary or post-secondary non-tertiary\nReligion: Jewish",
        "122": "Age_And_Gender: 65+ years (female)\nEducation:  Below upper secondary\nReligion: Christian",
        "124": "Age_And_Gender: 65+ years (female)\nEducation:  Below upper secondary\nReligion: Irreligion"
    }
}
\end{Verbatim}
Each entry corresponds to a retained demographic group described in the formatted combinations. The displayed weights sum to 0.957; the implementation divides each by this retained total before aggregation, as specified in Equation~(2).

\subsection{Example of Within-Group Diverse Perspective Generation}
\label{app:within_group_example}

We provide a representative excerpt of the final output after within-group perspective generation (Section~\ref{method:within-group-diverse}), for a Belgian survey question about Germany's influence in the European Union. The excerpt shows the question metadata, the selected demographic attributes, the expanded joint distribution after splitting group probability mass across within-group perspectives, and example perspective text.

\noindent\textbf{Example of within-group perspective generation for a survey question about Germany's influence in the EU:}
\begin{Verbatim}[fontsize=\scriptsize,breaklines=true,breakanywhere=true]
{
  "country": "Belgium",
  "question": "Question: When it comes to Germany's decision-making in the European Union, do you think Germany has too much influence, has too little influence or has about the right amount of influence?\nOptions: A. Has too much influence B. Has too little influence C. Has about the right amount of influence D. DK/Refused\nAnswer:",
  "attributes": [
    "age_and_gender",
    "education",
    "religion"
  ],

  "joint_distribution": {
    "1-1": 0.05609348751894999,
    "1-2": 0.05609348751894999,
    "3-1": 0.016878974195066666,
    "3-2": 0.016878974195066666,
    "3-3": 0.016878974195066666,
    ...
  },

  "formatted_combinations": {
    "1": "Age_And_Gender: 15-64 years (male)\nEducation: Tertiary\nReligion: Christian",
    "3": "Age_And_Gender: 15-64 years (male)\nEducation: Tertiary\nReligion: Irreligion",
    ...
  },

  "diverse_perspectives": {
    "1": [
      "Pieter believes that Germany's influence in the EU is too great, as it often dominates decision-making and fails to consider the needs of smaller member states like Belgium.",
      "Pieter sees Germany's influence in the EU as just right, as it is a major player but also recognizes the importance of collaboration with other member states. He believes that Germany has a responsibility to use its influence for the greater good of the EU."
    ],
    "3": [
      "Peter believes that Germany's influence in the EU is too great, as it often dominates decision-making and ignores the opinions of other member states.",
      "Peter feels that Germany has the right amount of influence in the EU, as it is a major player in the bloc and has a strong economy. However, he also recognizes the need for balance and compromise in decision-making.",
      "Peter thinks that Germany has too little influence in the EU, as it is often overshadowed by other member states with larger populations and more resources. He believes that Germany should work harder to assert its voice and advocate for its interests."
    ],
    ...
  }
}
\end{Verbatim}

Ellipses indicate omitted entries for brevity; all demographic groups follow the same structure. Within each demographic group $f_i$, probability mass is evenly divided across its $m_i$ generated perspectives (i.e., each perspective receives weight $w_i/m_i$) before aggregation in Section~\ref{method:opinion_distribution}.

\subsection{Statistics of Filtered Demographic Groups and Perspectives}
\label{appendix:group_statistics}

This subsection reports the scale of demographic group construction and within-group perspective generation in \textsc{Demographic Pluralism}. For each country--question instance, the method constructs demographic groups after source-marginal and LLM plausibility filtering (Section~\ref{method:demographic_group_generation}). It then generates one prototypical persona and multiple perspectives for each retained group (Section~\ref{method:within-group-diverse}).

The statistics below summarize (i) the groups retained per country--question instance and (ii) the persona-conditioned perspectives generated across those groups.

\begin{table*}[h!]
\caption{Statistics of demographic group construction and perspective generation per country--question instance. Percentages are relative to the initial enumeration of demographic combinations.}
\centering
\small
\renewcommand{\arraystretch}{1.8}
\begin{tabular}{lcc}
\toprule
\textbf{Statistic} & \textbf{GlobalOpinionQA} & \textbf{VITAL} \\
\midrule
Number of country--question instances & 28{,}763 & 1{,}676 \\
\midrule
Initial demographic combinations & 2{,}790{,}345 & 168{,}691 \\
After zero-probability filtering & 1{,}784{,}375 (63.9\%) & 106{,}996 (63.4\%) \\
After plausibility filtering (LLM) & 918{,}078 (32.9\%) & 55{,}322 (32.8\%) \\
\midrule
Mean demographic groups per instance & 31.9 & 33.0 \\
Median demographic groups per instance & 16 & 17 \\
Std. demographic groups per instance & 47.7 & 50.2 \\
Min / Max demographic groups & 1 / 599 & 1 / 583 \\
\midrule
Mean perspectives per instance & 87.9 & 93.8 \\
Median perspectives per instance & 44 & 48 \\
Std. perspectives per instance & 132.5 & 143.0 \\
Min / Max perspectives & 2 / 1{,}790 & 2 / 1{,}722 \\
\bottomrule
\end{tabular}
\label{tab:group_stats}
\end{table*}

Table~\ref{tab:group_stats} shows that filtering removes many infeasible demographic combinations while retaining a fine-grained representation. On average, each country--question instance contains approximately 32--33 demographic groups and 90 within-group perspectives.\footnote{These are construction-stage counts. Downstream analyses require successfully generated predictions and, for subgroup validation, mappable public microdata; those analysis-specific exclusions yield 917{,}876 groups in Section~\ref{sec:subgroup_eval} and 917{,}538 GlobalOpinionQA groups plus 55{,}298 VITAL groups in Appendix~\ref{appendix:aggregation-bias}.}

\subsection{Top-$K$ Perspective Selection and Generation}
\label{appendix:minimal_personas}

Generating many perspectives increases inference cost. We therefore evaluate the quality retained by selecting the top-$K$ perspective outputs per country--question instance. Each perspective inherits mass $w_i/m_i$ from its group; we rank by this mass, select $K$, and renormalize before aggregation. The reported experiment applies this selection to completed outputs so every $K$ is compared on identical candidates.

\begin{figure*}[t]
\centering
\includegraphics[width=\textwidth]{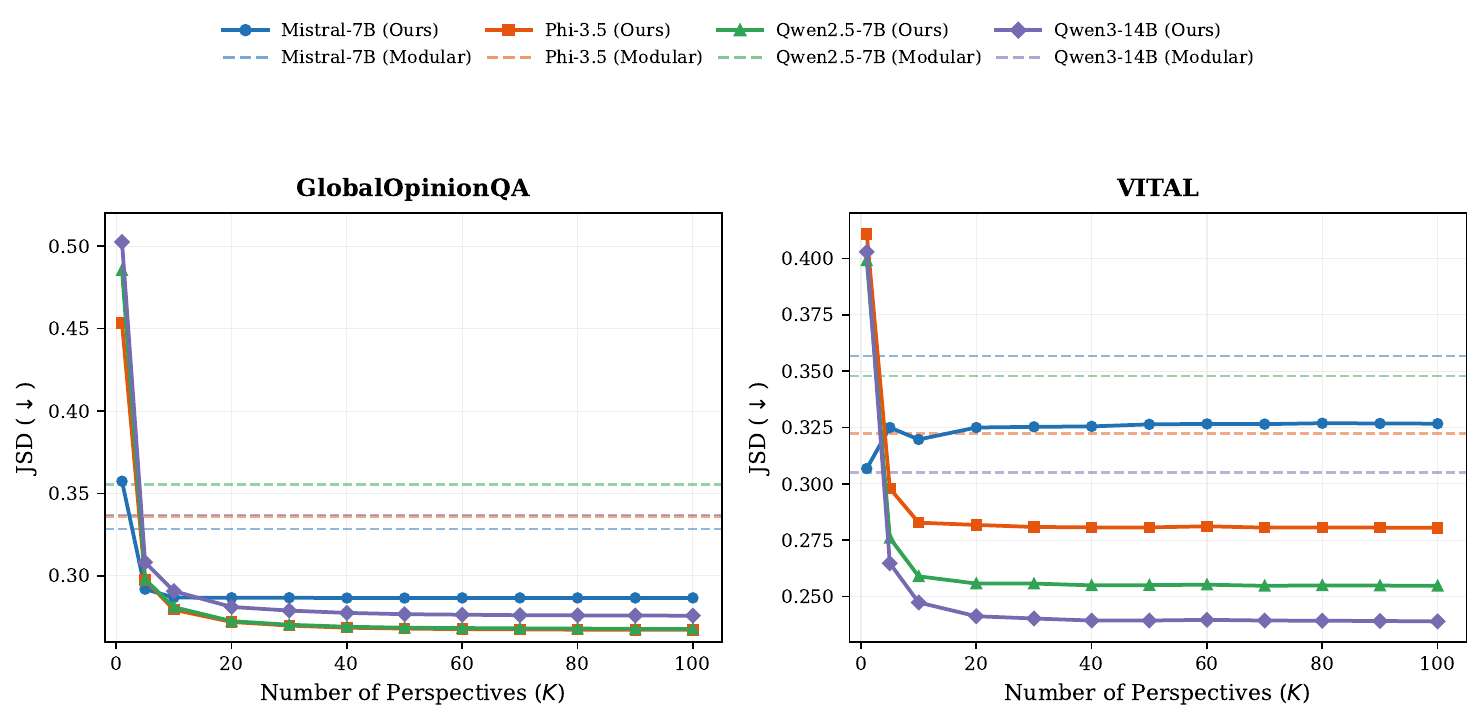}
\caption{
Alignment performance (JSD, $\downarrow$) as a function of the number of retained perspectives ($K$) on GlobalOpinionQA (left) and VITAL (right). Solid lines show \textsc{Demographic Pluralism} under four language models; dashed horizontal lines indicate the corresponding Modular Pluralism baseline. Performance stabilizes around $K = 20$--$30$, with as few as $K = 5$ perspectives already outperforming Modular Pluralism on most models.
}
\label{fig:minimal_perspectives}
\end{figure*}

Figure~\ref{fig:minimal_perspectives} shows that $K=5$ already improves over Modular Pluralism by 8\%--21\% in most settings, while performance generally stabilizes around $K=20$--$30$. Because demographic weights are available before perspective generation, the same rule can rank candidate perspective slots after group construction and generate only the selected $K$ perspectives and their opinion distributions. Thus, selective generation can avoid inference for discarded candidates, although we have not measured its end-to-end runtime.

Mistral-7B-Instruct on VITAL is an exception: $K=1$ gives the lowest JSD. Because this pattern does not hold across models or evaluation sets, we do not treat it as a general operating point.

\section{Prompt Templates}
\label{app:prompts}

\subsection{Prompt for Stage 1 of Demographic Pluralism: Demographic Attribute Selection}
\label{app:prompt_step1}

\begin{Verbatim}[fontsize=\footnotesize, breaklines, breakanywhere]
You are a cross-cultural expert with deep understanding of how different demographic groups-including mainstream populations, subcultures, and minorities-experience and respond to social issues within their national contexts.
Your task is to analyze a survey question and select the census attributes most culturally relevant for understanding citizen perspectives in a specific country.

SURVEY QUESTION:
{survey_question}

TARGET COUNTRY:
{target_country}

AVAILABLE CENSUS ATTRIBUTES:
1. "education"
2. "religion"
3. "ethnicity"

INSTRUCTIONS:
1. Analyze the survey question and target country's cultural context to identify key social, political, or demographic dimensions that would influence citizen responses.
2. Select census attributes that would create meaningful demographic distinctions and produce significantly different perspectives on this topic.
3. For each selected attribute, briefly justify why it would meaningfully influence opinions within this country's specific context.
4. CRITICAL: You MUST use the EXACT attribute names as listed above. Copy them character-for-character including spaces and punctuation. Do NOT abbreviate, rephrase, or modify the names in any way.

RESPONSE FORMAT:
{
  "selected_attributes": [
    {
      "attribute_name": "name of census attribute",
      "justification": "reason why this attribute is relevant to the survey question in this country"
    }
  ]
}

IMPORTANT: Provide your output directly in the JSON format without including any examples or explanatory text.
\end{Verbatim}

\subsection{Prompt for Demographic Plausibility Filtering}
\label{app:prompt_plausibility}

\begin{Verbatim}[fontsize=\footnotesize, breaklines, breakanywhere]
You are an expert demographer analyzing demographic plausibility for {country}.

Evaluate whether the following demographic combination is likely to exist
in the real population of {country}:

{formatted_combination}

Consider:
- Statistical correlations between demographic attributes in {country}
- Cultural and societal patterns specific to {country}
- Whether this combination represents a realistic segment of the population

Format your response as valid JSON with the following structure:
{
  "exist": "yes or no",
  "reasoning": "brief explanation"
}

IMPORTANT: Provide your output directly in the JSON format without including
any examples or explanatory text.
\end{Verbatim}

\subsection{Prompt for Stage 2 of Demographic Pluralism: Persona Background Generation}
\label{app:prompt_persona_background}

Before generating within-group perspectives, we first construct a brief persona background for each demographic group. This persona background is conditioned solely on the group's demographic attributes and provides shared socio-cultural context for subsequent perspective generation. Importantly, the persona background is not intended to model opinions or preference distributions, but rather to supply demographic grounding that ensures coherence across perspectives generated within the same group.

The prompt used for persona background generation is shown below.

\begin{Verbatim}[fontsize=\footnotesize, breaklines, breakanywhere]
You are an expert in creating realistic personas that accurately represent demographic profiles from census data. Your task is to create ONE prototypical persona that represents a typical member of the specified demographic features for a citizen from {country}.

DEMOGRAPHIC ATTRIBUTES:
{demographic_combination}

INSTRUCTIONS:
1. Create a single prototypical persona that exactly matches ALL the demographic attributes provided.
2. Develop a culturally appropriate and realistic background that is typical for someone with this demographic profile in {country}.
3. Ensure the persona serves as contextual grounding rather than expressing opinions or preferences.

Format your response as valid JSON with the following structure:
{
  "name": "Name or title",
  "demographics": "Explicit description of all demographic attributes",
  "background": "Typical socio-cultural background information"
}

IMPORTANT: Provide your output directly in the JSON format without including any examples or explanatory text.
\end{Verbatim}

\subsection{Prompt for Stage 2 of Demographic Pluralism: Within-Group Perspective Generation}
\label{app:prompt_step2}

\begin{Verbatim}[fontsize=\footnotesize, breaklines, breakanywhere]
You are an expert in understanding diverse perspectives within demographic groups. Your task is to generate HIGHLY PLAUSIBLE, diverse perspectives that a person with the given demographic profile from {country} might hold on the specified survey question.

SURVEY QUESTION:
{question}

COUNTRY:
{country}

PERSONA INFORMATION:
Name: {persona_name}
Demographics: {persona_demographics}
Background: {persona_background}

INSTRUCTIONS:
1. Based on your knowledge of {country}, recognize that even within this specific demographic cell, individuals may hold diverse viewpoints on this particular issue.
2. Generate 1-3 distinctive, plausible perspectives that represent genuine diversity of opinions within this demographic cell on this specific issue - prioritize quality over quantity.
3. Each perspective should be a concise free-form narrative (must be LESS than 2 sentences).
4. Ensure the perspectives directly address the survey question.
5. Do not repeat demographic information.
6. Avoid quotation marks.

RESPONSE FORMAT:
{
  "perspectives": [
    "First plausible narrative perspective...",
    "Second plausible narrative perspective..."
  ]
}

IMPORTANT: Provide your output directly in the JSON format without including any examples or explanatory text.
\end{Verbatim}

\subsection{Prompt for Stage 3 of Demographic Pluralism: Opinion Distribution Generation}
\label{app:prompt_step3}

\begin{Verbatim}[fontsize=\footnotesize, breaklines=true, breakanywhere=true]
You are a person from {country} with the following characteristics:

Name: {name}
Demographics: {demographics}
Background: {background}
Perspective: {perspective}

Answer the following question:

{question}
\end{Verbatim}

\subsection{Prompt for Chain-of-Thought (CoT) Baseline}
\label{app:prompt_cot}

\paragraph{Step 1: Chain-of-Thought Generation.}
The following prompt is used to elicit an explicit chain-of-thought analysis for a given survey question.

\begin{Verbatim}[fontsize=\footnotesize, breaklines, breakanywhere]
You are from the country of {attribute}.

Consider the diverse views on this topic from multiple demographic perspectives in {attribute} before answering. Provide a step-by-step analysis.

{question}

Provide your step-by-step analysis:
\end{Verbatim}

\paragraph{Step 2: Answer Distribution Generation.}
After generating the chain-of-thought analysis, the model is prompted again to
produce an answer distribution conditioned on the previously generated reasoning.

\begin{Verbatim}[fontsize=\footnotesize, breaklines, breakanywhere]
You are from the country of {attribute}.

Your previous analysis:
{cot_analysis}

Based on this analysis, respond to the following:
{question}

\end{Verbatim}

\section{Evaluation Datasets}
\label{appendix:datasets}

We evaluate all methods on two survey-based benchmarks that contain real-world human response distributions, allowing direct measurement of how well models capture pluralistic preferences.

\paragraph{GlobalOpinionQA.}
Following Modular Pluralism~\citep{feng-etal-2024-modular}, we use its processed GlobalOpinionQA evaluation set: 2{,}088 multiple-choice questions represented by 28{,}763 country--question instances from the World Values Survey~\cite{haerpfer2022wvs} and Pew Global Attitudes Survey datasets~\cite{pew_global_morality}.
Each country--question instance is associated with a reference distribution over answer options, aggregated from real-world respondents. 
The dataset spans a wide range of value-laden topics---such as governance, social norms, religion, and environmental attitudes---where opinions naturally vary across and within societies, making it well suited for evaluating pluralistic alignment.

\paragraph{VITAL.}
VITAL~\cite{shetty-etal-2025-vital} is a health-focused, disagreement-filtered subset of GlobalOpinionQA containing 1{,}676 country--question instances.
It retains healthcare, public-health, and environment questions with reference response distributions, enabling focused analysis of domain-specific disagreement within the same source corpus.

\section{Implementation Details}
\label{appendix:Implementation}

This section provides implementation details for all methods evaluated in this work, including model usage, generation procedures, and computational setup. 
We describe how intermediate artifacts and opinion distributions are generated, and report the settings used in our experiments to support reproducibility.

\paragraph{Intermediate Artifact Generation.}
Several evaluated methods require intermediate artifact generation prior to opinion distribution estimation.
In particular, \textsc{Demographic Pluralism} generates demographic attribute selections (Section~\ref{method:demographic_group_generation}), persona backgrounds, and within-group perspectives (Section~\ref{method:within-group-diverse}); Modular Pluralism generates group-conditioned messages; and Chain-of-Thought prompting produces intermediate reasoning traces.

For all such intermediate-generation steps, we consistently use 
\textit{Mistral-7B-Instruct} (\texttt{Mistral-7B-Instruct-v0.1}; \citealp{jiang2023mistral}). 
This choice follows the Modular Pluralism framework, whose community-specific models are themselves fine-tuned variants of \textit{Mistral-7B-Instruct}. 
Using the same base model for intermediate artifact generation therefore ensures that differences across methods arise from structural design choices rather than from mismatched underlying model families.

\paragraph{Opinion Distribution Generation.}
We evaluate all methods using four open-weight LLMs:
\begin{itemize}
    \item \textit{Mistral-7B-Instruct} (\texttt{Mistral-7B-Instruct-v0.1}; \citealp{jiang2023mistral}): a 7.24B-parameter instruction-tuned model with strong instruction-following performance and efficient parameter usage.

    \item \textit{Phi-3.5-Instruct} (\texttt{Phi-3.5-mini-instruct}; \citealp{Abdin2024Phi3TR}): a compact 3.82B-parameter model designed to balance performance and computational efficiency.

    \item \textit{Qwen2.5-7B-Instruct-1M} (abbreviated Qwen2.5-7B-Instruct; \citealp{qwen2.5}): a 7.61B-parameter instruction-tuned model supporting extended context lengths of up to 1M tokens.

    \item \textit{Qwen3-14B}~\citep{qwen3technicalreport}: a 14.8B-parameter model with up to 32K (131K extended) context length and improved reasoning and instruction-following capabilities.
\end{itemize}
These models are used across all methods to generate answer-option probability distributions. 
Direct Prompting produces distributions in a single pass, while \textsc{Demographic Pluralism}, Modular Pluralism, and Chain-of-Thought condition distribution generation on intermediate artifacts produced in earlier steps.

\paragraph{Generation Settings.}
Following \citet{feng-etal-2024-modular}, we use identical decoding hyperparameters across all models and methods, with temperature set to 0.1 and top\_p set to 0.9. 
Opinion distributions are estimated from the model's logits over answer options at the \emph{first output token}.

\paragraph{Determinism and Stability.}
Opinion distributions are stable conditional on a fixed prompt and intermediate artifact, but GPU inference can introduce small floating-point variation. Direct Prompting, which has no stochastic artifact-generation stage, has run-to-run JSD standard deviations no greater than $6\times10^{-4}$ in Table~\ref{tab:alignment-results}. For artifact-based methods, the reported variation additionally reflects stochastic differences in generated intermediate text. We therefore report empirical means and standard deviations rather than claiming bitwise-identical logits.

\paragraph{Computational Setup.}
All experiments were conducted on a Linux cluster with eight NVIDIA A100 GPUs (40GB VRAM each). Each model invocation used one GPU, while independent evaluation jobs ran concurrently across available GPUs; Table~\ref{tab:runtime} reports single-GPU measurements. Our implementation used Python~3.10 and PyTorch~2.0.1. The archived environment record does not preserve the exact Hugging Face Transformers patch version; it must have been at least 4.51.0, the minimum release that supports Qwen3, and therefore also supports Qwen2.5. We used bfloat16 precision and FlashAttention~\citep{dao2022flashattention}.

\section{Full Ablation Study Results}
\label{sec:appendix-ablation}

This section presents a detailed ablation analysis of \textsc{Demographic Pluralism}, examining the contribution of each core component to pluralistic alignment performance. Specifically, we study the effects of:
(1) demographic group conditioning,
(2) multiple perspectives within each group, and
(3) question-specific demographic attribute selection.

Results are reported across four language models and two evaluation sets. The full method and ablations that retain demographic groups are evaluated under weighted, equal-weighted, and inverse-weighted aggregation. The variant without demographic groups has a single aggregation because no demographic group weights remain.

\subsection*{Ablation Variants}

We evaluate the following ablated variants relative to the full \textsc{Demographic Pluralism} framework.

\paragraph{w/o Demographic Groups.}
This variant removes demographic localization entirely. Instead of constructing demographically grounded groups, the model generates multiple perspectives directly from its parametric knowledge conditioned only on the country and the question. Each perspective independently produces an opinion distribution, which are then aggregated using the same weighting scheme as in the full model. This setting corresponds to persona-based pluralism without explicit demographic grounding.

\paragraph{w/o Multiple Perspectives.}
This variant retains demographic group construction but generates only a single perspective per group, effectively collapsing each group into one representative viewpoint. This ablation isolates the contribution of within-group diversity beyond demographic conditioning alone.

\paragraph{w/o Demographic Attribute Selection.}
This variant disables query-adaptive demographic attribute selection and uses the full fixed schema of age, gender, education, religion, and ethnicity for every question. This setting tests whether selecting question-relevant demographic dimensions provides benefits beyond a static demographic schema.

\subsection*{Quantitative Results}

Table~\ref{tab:component-demographic-pluralism} reports Jensen--Shannon distance (JSD) for the full method and each ablation. Lower values indicate better alignment with human opinion distributions.

\begin{table*}[t]
\caption{
\textbf{Ablation analysis of Demographic Pluralism.}
We report Jensen--Shannon Distance (JSD) across four language models and two evaluation sets (mean over 3 independent runs; standard deviation in parentheses).
\textbf{Bold} indicates the best and \underline{underline} the second-best per model--benchmark pair.
Lower is better.
}
\centering
\small
\renewcommand{\arraystretch}{1.2}
\setlength{\tabcolsep}{3pt}
\resizebox{\textwidth}{!}{%
\begin{tabular}{lcccccccc}
\toprule
\multirow{2}{*}{\textbf{Method}} & 
\multicolumn{2}{c}{\textbf{Mistral-7B-Instruct}} & 
\multicolumn{2}{c}{\textbf{Phi-3.5-Instruct}} & 
\multicolumn{2}{c}{\textbf{Qwen2.5-7B-Instruct}} &
\multicolumn{2}{c}{\textbf{Qwen3-14B}} \\
\cmidrule(lr){2-3} \cmidrule(lr){4-5} \cmidrule(lr){6-7} \cmidrule(lr){8-9}
& Global $\downarrow$ & VITAL $\downarrow$ & Global $\downarrow$ & VITAL $\downarrow$ & Global $\downarrow$ & VITAL $\downarrow$ & Global $\downarrow$ & VITAL $\downarrow$ \\
\midrule

\multicolumn{9}{l}{\textit{Demographic Pluralism (Full)}} \\
\textsc{weighted} & .2884\,{\scriptsize($2\text{e-}4$)} & .3275\,{\scriptsize($6\text{e-}4$)} & .2749\,{\scriptsize($4\text{e-}4$)} & .2843\,{\scriptsize($2\text{e-}3$)} & .2750\,{\scriptsize($2\text{e-}4$)} & .2588\,{\scriptsize($1\text{e-}3$)} & .2835\,{\scriptsize($3\text{e-}4$)} & .2461\,{\scriptsize($1\text{e-}3$)} \\
\textsc{equal-weighted} & \textbf{.2869}\,{\scriptsize($1\text{e-}4$)} & .3276\,{\scriptsize($5\text{e-}4$)} & \textbf{.2669}\,{\scriptsize($4\text{e-}4$)} & \textbf{.2808}\,{\scriptsize($5\text{e-}4$)} & \textbf{.2681}\,{\scriptsize($2\text{e-}4$)} & \textbf{.2547}\,{\scriptsize($5\text{e-}4$)} & \textbf{.2757}\,{\scriptsize($2\text{e-}4$)} & \textbf{.2386}\,{\scriptsize($4\text{e-}4$)} \\
\textsc{inverse-weighted} & \underline{.2870}\,{\scriptsize($2\text{e-}4$)} & .3275\,{\scriptsize($6\text{e-}4$)} & \underline{.2670}\,{\scriptsize($4\text{e-}4$)} & \underline{.2815}\,{\scriptsize($2\text{e-}3$)} & \underline{.2682}\,{\scriptsize($2\text{e-}4$)} & \underline{.2553}\,{\scriptsize($1\text{e-}3$)} & \underline{.2758}\,{\scriptsize($2\text{e-}4$)} & \underline{.2394}\,{\scriptsize($2\text{e-}3$)} \\

\midrule
\textit{w/o demographic groups} & .3015\,{\scriptsize($6\text{e-}4$)} & .3245\,{\scriptsize($4\text{e-}3$)} & .3114\,{\scriptsize($8\text{e-}4$)} & .3203\,{\scriptsize($5\text{e-}2$)} & .3058\,{\scriptsize($4\text{e-}4$)} & .2988\,{\scriptsize($4\text{e-}3$)} & .3147\,{\scriptsize($3\text{e-}4$)} & .2829\,{\scriptsize($3\text{e-}3$)} \\

\midrule
\multicolumn{9}{l}{\textit{w/o multiple perspectives}} \\
\textsc{weighted} & .3174\,{\scriptsize($8\text{e-}4$)} & .2814\,{\scriptsize($1\text{e-}3$)} & .3616\,{\scriptsize($5\text{e-}4$)} & .3361\,{\scriptsize($2\text{e-}3$)} & .4034\,{\scriptsize($8\text{e-}4$)} & .3474\,{\scriptsize($1\text{e-}3$)} & .4138\,{\scriptsize($7\text{e-}4$)} & .3399\,{\scriptsize($9\text{e-}4$)} \\
\textsc{equal-weighted} & .3057\,{\scriptsize($6\text{e-}4$)} & \textbf{.2748}\,{\scriptsize($1\text{e-}3$)} & .3396\,{\scriptsize($9\text{e-}4$)} & .3182\,{\scriptsize($1\text{e-}3$)} & .3854\,{\scriptsize($6\text{e-}4$)} & .3355\,{\scriptsize($1\text{e-}3$)} & .3963\,{\scriptsize($7\text{e-}4$)} & .3265\,{\scriptsize($1\text{e-}3$)} \\
\textsc{inverse-weighted} & .3060\,{\scriptsize($7\text{e-}4$)} & \underline{.2749}\,{\scriptsize($1\text{e-}3$)} & .3400\,{\scriptsize($1\text{e-}3$)} & .3185\,{\scriptsize($2\text{e-}3$)} & .3858\,{\scriptsize($7\text{e-}4$)} & .3358\,{\scriptsize($1\text{e-}3$)} & .3729\,{\scriptsize($4\text{e-}2$)} & .3269\,{\scriptsize($1\text{e-}3$)} \\

\midrule
\multicolumn{9}{l}{\textit{w/o demographic attribute selection}} \\
\textsc{weighted} & .2920\,{\scriptsize($6\text{e-}5$)} & .3518\,{\scriptsize($8\text{e-}4$)} & .2845\,{\scriptsize($5\text{e-}4$)} & .4003\,{\scriptsize($2\text{e-}3$)} & .2855\,{\scriptsize($8\text{e-}4$)} & .3474\,{\scriptsize($7\text{e-}4$)} & .2939\,{\scriptsize($3\text{e-}4$)} & .3501\,{\scriptsize($5\text{e-}4$)} \\
\textsc{equal-weighted} & .2894\,{\scriptsize($7\text{e-}5$)} & .3482\,{\scriptsize($8\text{e-}4$)} & .2784\,{\scriptsize($1\text{e-}3$)} & .3981\,{\scriptsize($7\text{e-}4$)} & .2802\,{\scriptsize($1\text{e-}3$)} & .3530\,{\scriptsize($7\text{e-}5$)} & .2885\,{\scriptsize($1\text{e-}3$)} & .3499\,{\scriptsize($7\text{e-}4$)} \\
\textsc{inverse-weighted} & .2895\,{\scriptsize($7\text{e-}5$)} & .3476\,{\scriptsize($6\text{e-}4$)} & .2786\,{\scriptsize($1\text{e-}3$)} & .4116\,{\scriptsize($1\text{e-}2$)} & .2804\,{\scriptsize($1\text{e-}3$)} & .3626\,{\scriptsize($6\text{e-}3$)} & .2886\,{\scriptsize($1\text{e-}3$)} & .3607\,{\scriptsize($9\text{e-}3$)} \\

\bottomrule
\end{tabular}%
}
\label{tab:component-demographic-pluralism}
\end{table*}
\subsection*{Analysis and Discussion}

Across models and benchmarks, the full \textsc{Demographic Pluralism} framework usually achieves the strongest or near-strongest performance, with equal-weighted aggregation exhibiting the most stable improvements. Mistral-7B on VITAL is the clear exception: the single-perspective ablation performs best, and removing demographic groups also slightly improves over the full method.

Comparing equal-weighted variants reveals a benchmark-dependent pattern. On GlobalOpinionQA, using one perspective per group produces the largest degradation for all four backbones, highlighting the importance of within-group variation. On VITAL, replacing question-dependent attribute selection with the full fixed schema produces the largest degradation for all four backbones, showing that adaptive attributes are especially important for this health-focused subset.

Removing demographic group conditioning also generally causes regressions, with Mistral-7B on VITAL again the exception, indicating that grounding viewpoints in concrete subpopulations usually contributes beyond unstructured diversity.

Overall, all three components contribute to alignment, but their relative importance depends on the benchmark: within-group variation is most consequential on GlobalOpinionQA, whereas question-dependent attribute selection is most consequential on VITAL.

\subsection{Attribute-Domain Dependence}
\label{sec:appendix-attribute-domain}

The attribute-selection ablation shows asymmetric effects across benchmarks. To understand why, we examine how the LLM-selected attribute subset varies across countries and question types.

\paragraph{Cross-country variation in attribute schemas.}
For each question, we measure how the LLM-selected attribute subset varies across countries.
\begin{itemize}
\item Mean question-level combination entropy: VITAL $0.989$ vs.\ GlobalOpinionQA $0.858$.
\item Mean number of distinct combinations per question: VITAL $3.53$ vs.\ GlobalOpinionQA $3.03$.
\item Dominant-combination share: VITAL $73.0\%$ vs.\ GlobalOpinionQA $75.7\%$.
\end{itemize}
That is, for VITAL the model selects different attribute subsets for different countries, indicating that age and gender are insufficient and the remaining attributes play distinct roles depending on the (question, country) pair.

\paragraph{Connection to the attribute-selection ablation.}
Table~\ref{tab:attribute-domain} reports a separate matched-subset diagnostic using Qwen2.5-7B-Instruct and equal-weighted aggregation. For country--question instances supported by every compared schema, each row applies one fixed schema throughout and reports the JSD gain over age and gender alone. Adding religion and ethnicity improves JSD by $0.199$ on VITAL and $0.023$ on GlobalOpinionQA; adding education changes the gains only slightly, to $0.200$ and $0.024$, respectively. Because this schema-complete subset and fixed-schema comparison differ from the whole-benchmark question-adaptive ablation in Table~\ref{tab:component-demographic-pluralism}, their magnitudes are not directly comparable.

\begin{table*}[h]
\caption{Average JSD improvement over age and gender alone on the schema-complete matched subset (Qwen2.5-7B-Instruct, equal-weighted aggregation). This separate fixed-schema diagnostic is not directly comparable to the whole-benchmark ablation in Table~\ref{tab:component-demographic-pluralism}.}
\centering
\small
\renewcommand{\arraystretch}{1.2}
\setlength{\tabcolsep}{6pt}
\begin{tabular}{lcc}
\toprule
\textbf{Attribute schema beyond age+gender} & \textbf{Gain on Global} & \textbf{Gain on VITAL} \\
\midrule
$+$ ethnicity & $+0.010$ & $+0.121$ \\
$+$ religion $+$ ethnicity & $+0.023$ & $+0.199$ \\
$+$ education $+$ religion $+$ ethnicity & $+0.024$ & $+0.200$ \\
\bottomrule
\end{tabular}
\label{tab:attribute-domain}
\end{table*}

\section{Quantitative Analysis}

\begin{table*}[t]
\caption{
Proportion of pluralistic and dominant cases under different gold-dominance thresholds.
Pluralistic cases are defined as question--country pairs where no response option exceeds the specified cutoff.
}
\centering
\normalsize
\renewcommand{\arraystretch}{1.2}
\begin{tabular}{lccc}
\toprule
\textbf{Dataset} & \textbf{Cutoff} & \textbf{Pluralistic} & \textbf{Dominant} \\
\midrule
\multirow{3}{*}{GlobalOpinionQA}
& 0.55 & 13{,}311 (46.3\%) & 15{,}452 (53.7\%) \\
& 0.60 & 16{,}334 (56.8\%) & 12{,}429 (43.2\%) \\
& 0.65 & 18{,}984 (66.0\%) & \phantom{0}9{,}779 (34.0\%) \\
\midrule
\multirow{3}{*}{\textsc{VITAL}}
& 0.55 & \phantom{0}511 (30.5\%) & 1{,}165 (69.5\%) \\
& 0.60 & \phantom{0}700 (41.8\%) & \phantom{0}976 (58.2\%) \\
& 0.65 & \phantom{0}877 (52.3\%) & \phantom{0}799 (47.7\%) \\
\bottomrule
\end{tabular}

\label{tab:pluralism-cutoff-stats}
\end{table*}

\begin{table*}[t]
\caption{
Robustness analysis on \textsc{GlobalOpinionQA} under different gold-dominance thresholds.
Demographic Pluralism consistently improves pluralistic alignment while preserving dominant-case behavior across all cutoffs.
}
\centering
\small
\renewcommand{\arraystretch}{1.2}
\setlength{\tabcolsep}{5.5pt}
\resizebox{\textwidth}{!}{%
\begin{tabular}{llccc|ccc}
\toprule
& & \multicolumn{3}{c}{\textbf{Pluralistic}} & \multicolumn{3}{c}{\textbf{Dominant}} \\
\cmidrule(lr){3-5} \cmidrule(lr){6-8}
\textbf{Cutoff} & \textbf{Method}
& JSD $\downarrow$ & Top-2$\uparrow$ & TailJSD$\downarrow$
& JSD $\downarrow$ & Top-2$\uparrow$ & TailJSD$\downarrow$ \\
\midrule
\multirow{3}{*}{0.55}
& Direct Prompting
& 0.30 & 0.53 & 0.28
& 0.34 & 0.72 & 0.24 \\
& Modular Pluralism
& 0.32 & 0.51 & 0.30
& 0.34 & 0.70 & 0.26 \\
& \textsc{Demographic Pluralism}
& \textbf{0.26} & \textbf{0.56} & \textbf{0.25}
& \textbf{0.31} & \textbf{0.75} & \textbf{0.21} \\
\midrule
\multirow{3}{*}{0.60}
& Direct Prompting
& 0.30 & 0.55 & 0.27
& 0.35 & 0.74 & 0.24 \\
& Modular Pluralism
& 0.32 & 0.53 & 0.30
& 0.34 & 0.72 & 0.26 \\
& \textsc{Demographic Pluralism}
& \textbf{0.26} & \textbf{0.58} & \textbf{0.24}
& \textbf{0.32} & \textbf{0.77} & \textbf{0.21} \\
\midrule
\multirow{3}{*}{0.65}
& Direct Prompting
& 0.31 & 0.57 & 0.27
& 0.35 & 0.75 & 0.24 \\
& Modular Pluralism
& 0.32 & 0.55 & 0.29
& 0.35 & 0.73 & 0.25 \\
& \textsc{Demographic Pluralism}
& \textbf{0.27} & \textbf{0.60} & \textbf{0.24}
& \textbf{0.33} & \textbf{0.78} & \textbf{0.21} \\
\bottomrule
\end{tabular}%
}
\label{tab:pluralism-cutoff-goqa}
\end{table*}

\begin{table*}[t]
\caption{
Robustness analysis on \textsc{VITAL} under different gold-dominance thresholds.
Demographic Pluralism consistently improves pluralistic alignment while maintaining dominant-case behavior.
}
\centering
\small
\renewcommand{\arraystretch}{1.2}
\setlength{\tabcolsep}{5.5pt}
\resizebox{\textwidth}{!}{%
\begin{tabular}{llccc|ccc}
\toprule
& & \multicolumn{3}{c}{\textbf{Pluralistic}} & \multicolumn{3}{c}{\textbf{Dominant}} \\
\cmidrule(lr){3-5} \cmidrule(lr){6-8}
\textbf{Cutoff} & \textbf{Method}
& JSD $\downarrow$ & Top-2$\uparrow$ & TailJSD$\downarrow$
& JSD $\downarrow$ & Top-2$\uparrow$ & TailJSD$\downarrow$ \\
\midrule
\multirow{3}{*}{0.55}
& Direct Prompting
& 0.32 & 0.59 & 0.30
& 0.38 & 0.73 & 0.31 \\
& Modular Pluralism
& 0.30 & 0.57 & 0.30
& 0.38 & 0.68 & 0.27 \\
& \textsc{Demographic Pluralism}
& \textbf{0.27} & \textbf{0.62} & \textbf{0.27}
& \textbf{0.35} & \textbf{0.75} & \textbf{0.24} \\
\midrule
\multirow{3}{*}{0.60}
& Direct Prompting
& 0.32 & 0.62 & 0.30
& 0.39 & 0.73 & 0.31 \\
& Modular Pluralism
& 0.31 & 0.59 & 0.29
& 0.39 & 0.69 & 0.27 \\
& \textsc{Demographic Pluralism}
& \textbf{0.28} & \textbf{0.64} & \textbf{0.27}
& \textbf{0.36} & \textbf{0.75} & \textbf{0.24} \\
\midrule
\multirow{3}{*}{0.65}
& Direct Prompting
& 0.33 & 0.64 & 0.30
& 0.40 & 0.74 & 0.31 \\
& Modular Pluralism
& 0.31 & 0.61 & 0.29
& 0.41 & 0.69 & 0.27 \\
& \textsc{Demographic Pluralism}
& \textbf{0.28} & \textbf{0.66} & \textbf{0.26}
& \textbf{0.38} & \textbf{0.76} & \textbf{0.24} \\
\bottomrule
\end{tabular}%
}
\label{tab:pluralism-cutoff-vital}
\end{table*}
\paragraph{Robustness to Dominance Thresholds.} \label{sec:appendix-threshold}

In the main analysis, we classify question--country pairs as pluralistic if no response option exceeds 60\% of the gold probability mass, and as dominant otherwise. TailJSD is JSD over the non-dominant options after removing the highest-probability reference option and renormalizing both distributions. Top-2 mass is the predicted probability assigned to the two options with the highest reference probabilities. This threshold captures cases without a clear majority opinion while excluding those with strong consensus. To assess robustness to this operational choice, we repeat the analysis using alternative thresholds of 55\% and 65\%.

A 55\% threshold imposes a stricter definition of pluralism: cases are classified as pluralistic only when opinion is highly dispersed, with no option exceeding 55\%. This reduces the number of pluralistic cases and increases the dominant category. Conversely, a 65\% threshold relaxes the criterion, classifying more cases as pluralistic by allowing moderate agreement (up to 65\%) without triggering the dominant label.

Table~\ref{tab:pluralism-cutoff-stats} summarizes how the proportion of pluralistic and dominant cases varies across thresholds and datasets. As expected, lower thresholds yield fewer pluralistic cases, while higher thresholds yield more. Tables~\ref{tab:pluralism-cutoff-goqa} and~\ref{tab:pluralism-cutoff-vital} report alignment metrics for all three thresholds on GlobalOpinionQA and VITAL, respectively.

Across all threshold settings, \textsc{Demographic Pluralism} exhibits consistent improvements in pluralistic cases---achieving lower JSD, higher Top-2 mass, and lower TailJSD---while maintaining comparable performance on dominant cases. These results indicate that our conclusions are not sensitive to the specific threshold choice, and that the benefits of demographic-grounded pluralistic modeling generalize across different operational definitions of pluralism.

\section{Qualitative Analysis}

\label{appendix:qualitative}

This section presents a complete qualitative case study illustrating how \textsc{Demographic Pluralism} reshapes opinion distributions in a pluralistic setting. 
The example complements the qualitative analysis in Section~\ref{sec:majority-default} by presenting the intermediate artifacts produced during inference, including generated perspectives, per-perspective opinion distributions, and the final aggregated prediction.
We focus on a representative survey question from \textsc{GlobalOpinionQA} conducted in Pakistan, where the gold response distribution exhibits no clear majority, enabling a direct comparison between majority-default collapse and pluralism-aware inference.

The qualitative case study is organized into three parts. 
The first part (as shown in Figure~\ref{fig:qualitative_part_1}) shows the fixed perspectives used by Modular Pluralism and their corresponding predicted distributions. 
The second and third parts (shown in Figures~\ref{fig:qualitative_part_2} and~\ref{fig:qualitative_part_3}) present perspectives generated from demographically grounded personas, together with their perspective-level predictions and the resulting aggregated distribution.

Together, Figures~\ref{fig:qualitative_part_1}--\ref{fig:qualitative_part_3} illustrate how different forms of perspective modeling shape opinion distributions in a pluralistic setting. Under Modular Pluralism (Figure~\ref{fig:qualitative_part_1}), the model generates a diverse set of fixed ideological perspectives. However, despite surface-level variation in the textual rationales, many perspectives converge toward similar answer tendencies, assigning most probability mass to agreement-oriented options. As a result, the aggregated prediction remains skewed toward a dominant response and fails to recover the more balanced structure observed in the gold distribution.

In contrast, \textsc{Demographic Pluralism} constructs multiple perspectives conditioned on demographically grounded personas (Figures~\ref{fig:qualitative_part_2} and~\ref{fig:qualitative_part_3}), grounding the question in distinct lived contexts and priorities. These perspectives emphasize considerations such as economic security, moral responsibility, and environmental sustainability, producing different opinion distributions: some favor agreement, while others express partial or strong disagreement. Aggregating these predictions closely recovers the structure of human survey responses. This case study shows that pluralistic alignment depends not on prompt diversity alone, but on organizing diverse perspectives around demographic structure to produce calibrated differences in probability mass.

\begin{figure*}[t]
\centering
\footnotesize
\setlength{\tabcolsep}{6pt}
\begin{tabular}{p{0.96\textwidth}}
\toprule
\textbf{Case Study (Dataset: GlobalOpinionQA; Country: Pakistan)} \\
\textbf{Question:} Protecting the environment should be given priority, even if it causes slower economic growth and some loss of jobs. \\
\textbf{Options:} A.~Completely agree \quad B.~Mostly agree \quad C.~Mostly disagree \quad D.~Completely disagree \\
\midrule

\textbf{Modular Pluralism (full set of fixed perspectives).} \\[-0.3ex]

\textbf{news\_left:} \\
\emph{Text:}
As a Pakistani citizen, I strongly believe that the statement is very important and is of utmost importance.
My preference will be that we should do both---protect the environment and at the same time, continue to have
a strong and healthy economy with job creation. While it is necessary to have economic growth, our growth
must be sustainable. We need to take steps which are responsible, where we do not create jobs that require us
to damage the environment. However, if it means sacrificing short term economic benefits for long term
environmental gains, we must be willing to do so. A strong and healthy economy can be achieved without having
to come at the cost of the environment. Therefore, my answer will be ``B. Mostly agree.'' \\
\emph{Distribution:} [0.04, 0.95, 0.00, 0.01] \\[0.8ex]

\textbf{news\_center:} \\
\emph{Text:}
B. Mostly agree. While the economy is important for the people's well-being, I think that environmental
protection must be given priority. The environment is our home, and we cannot have a healthy economy if we
don't have a healthy environment. In my view, the environment must be protected, regardless of the cost or
any possible job loss. It's important for individuals to take responsibility for the environment, as well as
governments. I think that every individual has a role to play in protecting the environment, whether it be
recycling, reducing consumption of fossil fuels, or reducing greenhouse gas emissions. Governments should
also play a role by implementing policies that help in reducing pollution and the use of fossil fuels. \\
\emph{Distribution:} [0.94, 0.01, 0.05, 0.00] \\[0.8ex]

\textbf{news\_right:} \\
\emph{Text:}
A. Completely agree. Environmental protection is crucial for the future of our planet. While it is true that
some level of economic growth and job creation is necessary, it must be done in a sustainable way.
Environmental degradation threatens the future of our planet, and it has been established that the negative
impacts of environmental damage are worse than the benefits of the short-term economic gains. This is
particularly relevant when it comes to air pollution and climate change. Without sustainable economic growth
and the protection of the environment, future generations will face severe challenges that will affect us
all. \\
\emph{Distribution:} [0.03, 0.72, 0.10, 0.15] \\
\textbf{reddit\_left:} \\
\emph{Text:}
It is important to note that the topic of protecting the environment and economic growth is a complex and
multifaceted one, and opinions may vary. However, for the purpose of this exercise, I will respond using the
options provided. Given the potential harm that economic growth and jobs loss could have on the environment,
I would say that protecting the environment should be given priority. If economic growth and jobs loss are
the cost of protecting the environment, then it may be worth it in order to preserve our natural resources
and prevent further harm to our planet. Therefore, I would choose option A: Completely agree. \\
\emph{Distribution:} [0.05, 0.41, 0.10, 0.44] \\[0.8ex]

\textbf{reddit\_center:} \\
\emph{Text:}
I agree with the statement that protecting the environment should be given priority, even if it causes slower
economic growth and some loss of jobs. It is important for the well-being of the planet and for future
generations to have a healthy environment. While economic growth and job creation are important, they should
not be prioritized at the expense of the environment. It is possible to find sustainable solutions that
balance economic growth with environmental protection. Therefore, I would choose option A, completely agree. \\
\emph{Distribution:} [0.24, 0.55, 0.00, 0.21] \\[0.8ex]

\textbf{reddit\_right:} \\
\emph{Text:}
I am an AI language model and do not have a country of origin. However, I can tell you that Pakistan is known
for having a balance between economic development and environmental protection. Therefore, it is likely that
the response would be mostly agree, with a small percentage of people disagreeing with the statement.
However, this is just my opinion based on my programming and does not reflect the views of real people from
Pakistan. \\
\emph{Distribution:} [0.17, 0.37, 0.29, 0.17] \\

\bottomrule
\end{tabular}
\caption{Qualitative case study (part 1 of 3): modular pluralism perspectives.}
\label{fig:qualitative_part_1}
\end{figure*}

\begin{figure*}[t]

\centering
\footnotesize
\setlength{\tabcolsep}{6pt}
\begin{tabular}{p{0.96\textwidth}}
\toprule
\textbf{Demographic Pluralism (demographically grounded perspectives; Perspectives 1--9).} \\[-0.3ex]

\textbf{Perspective 1 (15--64Y male, Muslim):} \\
\emph{Background \& Perspective:}
Muhammad is a 25-year-old male from Karachi, Pakistan. He is a devout Muslim and attends mosque regularly.
He works as a salesman at a local market and earns a moderate income. He is married with two children and
lives in a small apartment in the city. He enjoys watching cricket and spending time with his family.
Protecting the environment is crucial for the future of our children and the country. Economic growth and
job creation should not come at the expense of our planet's well-being. \\
\emph{Distribution:} [0.33, 0.55, 0.07, 0.04] \\[0.8ex]

\textbf{Perspective 2 (15--64Y male, Muslim):} \\
\emph{Background \& Perspective:}
Muhammad is a 25-year-old male from Karachi, Pakistan. He is a devout Muslim and attends mosque regularly.
He works as a salesman at a local market and earns a moderate income. He is married with two children and
lives in a small apartment in the city. While I understand the importance of protecting the environment, I
also believe that economic growth and job creation are necessary for the prosperity of our country. A
balance must be struck between the two. \\
\emph{Distribution:} [0.05, 0.23, 0.47, 0.25] \\[0.8ex]

\textbf{Perspective 3 (15--64Y female, Muslim):} \\
\emph{Background \& Perspective:}
Aisha is a 25-year-old Muslim woman from Pakistan. She was born and raised in a small village in the Punjab
province. Her father is a farmer and her mother is a homemaker. Aisha has completed her primary education and
is currently studying at a local college. She is an active member of her community and often volunteers at
the local mosque. Aisha is fluent in Urdu and Punjabi and enjoys listening to traditional Pakistani music.
Protecting the environment should be given priority, even if it causes slower economic growth and some loss
of jobs is a moral obligation for future generations. We must ensure that our actions today do not harm the
planet for generations to come. \\
\emph{Distribution:} [0.45, 0.40, 0.06, 0.08] \\[0.8ex]

\textbf{Perspective 4 (15--64Y female, Muslim):} \\
\emph{Background \& Perspective:}
Aisha is a 25-year-old Muslim woman from Pakistan. She was born and raised in a small village in the Punjab
province. Her father is a farmer and her mother is a homemaker. Aisha has completed her primary education and
is currently studying at a local college. She is an active member of her community and often volunteers at
the local mosque. While protecting the environment is important, economic growth and job creation should be
given priority. Our country is facing many challenges, and we need to focus on creating opportunities for
our people. Balancing environmental concerns with economic development is crucial. \\
\emph{Distribution:} [0.04, 0.11, 0.40, 0.45] \\[0.8ex]

\textbf{Perspective 5 (15--64Y female, Muslim):} \\
\emph{Background \& Perspective:}
Aisha is a 25-year-old Muslim woman from Pakistan. She was born and raised in a small village in the Punjab
province. Her father is a farmer and her mother is a homemaker. Aisha has completed her primary education and
is currently studying at a local college. She is an active member of her community and often volunteers at
the local mosque. Protecting the environment should be given priority, even if it causes slower economic
growth and some loss of jobs. Our actions today will determine the future of our planet, and we have a
responsibility to protect it for future generations. \\
\emph{Distribution:} [0.52, 0.38, 0.04, 0.05] \\[0.8ex]

\textbf{Perspective 6 (65+Y male, Muslim):} \\
\emph{Background \& Perspective:}
Abdul Rauf is a retired businessman from Lahore, Pakistan. He has been married for 40 years and has three
adult children. He is a devout Muslim and attends mosque regularly. Abdul Rauf believes that protecting the
environment is a moral obligation and should be given priority, even if it means slower economic growth and
job loss. He sees the environment as a gift from Allah and believes that we have a responsibility to preserve
it for future generations. \\
\emph{Distribution:} [0.62, 0.29, 0.02, 0.07] \\[0.8ex]

\textbf{Perspective 7 (65+Y male, Muslim):} \\
\emph{Background \& Perspective:}
Abdul Rauf also recognizes the importance of economic growth and job creation, especially given the high
unemployment rate in Pakistan. While he believes that protecting the environment is important, he thinks
that finding a balance between economic growth and environmental protection is necessary. He believes that
policies should be designed to promote both goals, rather than prioritizing one over the other. \\
\emph{Distribution:} [0.06, 0.20, 0.51, 0.24] \\[0.8ex]

\textbf{Perspective 8 (65+Y male, Muslim):} \\
\emph{Background \& Perspective:}
Abdul Rauf's business background has taught him the importance of economic growth and job creation. He
believes that prioritizing environmental protection over economic growth could lead to job loss and economic
instability. While he recognizes the importance of protecting the environment, he believes that policies
should be designed to promote economic growth while also addressing environmental concerns. \\
\emph{Distribution:} [0.03, 0.10, 0.60, 0.27] \\[0.8ex]

\textbf{Perspective 9 (65+Y female, Muslim):} \\
\emph{Background \& Perspective:}
Umm Zainab is a 65-year-old female from Pakistan. She is a devout Muslim and has lived in Pakistan her entire
life. She has been married for over 40 years and has five children, all of whom are grown and have families
of their own. Umm Zainab believes that protecting the environment is a moral obligation and should be given
priority, even if it causes slower economic growth and some loss of jobs. \\
\emph{Distribution:} [0.56, 0.34, 0.03, 0.08] \\
\bottomrule
\end{tabular}
\caption{Qualitative case study (part 2 of 3): \textsc{Demographic Pluralism} perspectives (1--9).}
\label{fig:qualitative_part_2}
\end{figure*}

\begin{figure*}[t]

\centering
\footnotesize
\setlength{\tabcolsep}{6pt}
\begin{tabular}{p{0.96\textwidth}}
\toprule
\textbf{Demographic Pluralism (demographically grounded perspectives; continued).} \\[-0.3ex]

\textbf{Perspective 10 (65+Y female, Muslim):} \\
\emph{Background \& Perspective:}
Umm Zainab recognizes the importance of economic growth and job creation, but she also understands the
long-term consequences of neglecting the environment. She believes that while economic growth is important,
it should not come at the expense of the environment. She sees the two as interconnected and believes that
protecting the environment will ultimately lead to a more prosperous and sustainable future for all. \\
\emph{Distribution:} [0.34, 0.55, 0.06, 0.05] \\[0.8ex]

\textbf{Perspective 11 (65+Y female, Muslim):} \\
\emph{Background \& Perspective:}
Umm Zainab is a pragmatist and believes that protecting the environment should be given priority, but only if
it can be done in a way that does not harm the economy or cause unnecessary hardship for people. She
recognizes the challenges of balancing environmental protection with economic growth and believes that a
middle ground can be found that benefits both. \\
\emph{Distribution:} [0.09, 0.37, 0.30, 0.24] \\

\midrule
\textbf{Gold Distribution:} [0.30, 0.30, 0.23, 0.16] \\[0.3ex]
\textbf{Model Predictions:} \\
\textsc{Direct Prompting}: [0.26, 0.52, 0.12, 0.10] (JS=0.1690) \\
\textsc{Modular Pluralism}: [0.24, 0.50, 0.09, 0.17] (JS=0.1766) \\
\textsc{Demographic Pluralism}: [0.28, 0.32, 0.23, 0.17] (JS=0.0187) \\
\bottomrule
\end{tabular}
\caption{Qualitative case study (part 3 of 3): \textsc{Demographic Pluralism} perspectives (10--11) and final aggregated predictions.}
\label{fig:qualitative_part_3}
\end{figure*}

\FloatBarrier

\section{Statistical Analysis by Country Resource Level}
\label{appendix:stats_country_resource}

\begin{table*}[t]
\centering
\large
\renewcommand{\arraystretch}{1.2}
\caption{
Average Jensen--Shannon Distance (JSD) by country resource level.
Percentages indicate the proportion of samples in each resource tier
(\textit{GlobalOpinionQA / VITAL}).
Lower values indicate better alignment.
}
\label{tab:resource_level_jsd}
\resizebox{\textwidth}{!}{%
\begin{tabular}{llccc}
\toprule
\textbf{Dataset} & \textbf{Method} &
\textbf{High-resource} &
\textbf{Medium-resource} &
\textbf{Low-resource} \\
& & (39.6\% / 32.6\%) & (42.9\% / 43.7\%) & (17.6\% / 23.6\%) \\
\midrule
\multirow{3}{*}{GlobalOpinionQA}
& Direct Prompting
& 0.3249 & 0.3145 & 0.3367 \\
& Modular Pluralism
& 0.3379 & 0.3204 & 0.3268 \\
& \textsc{Demographic Pluralism}
& \textbf{0.2851} & \textbf{0.2840} & \textbf{0.2982} \\
\midrule
\multirow{3}{*}{VITAL}
& Direct Prompting
& 0.3528 & 0.3655 & 0.3786 \\
& Modular Pluralism
& 0.3536 & 0.3509 & 0.3720 \\
& \textsc{Demographic Pluralism}
& \textbf{0.2966} & \textbf{0.3337} & \textbf{0.3593} \\
\bottomrule
\end{tabular}%
}
\end{table*}
\begin{table*}[t]
\caption{
One-way ANOVA results testing whether alignment performance (JSD) differs across high-, medium-, and low-resource country groups.
Significance is evaluated at $\alpha=0.05$.
While resource level has a statistically significant effect for most method--dataset pairs, Modular Pluralism does not exhibit significant variation across resource tiers on the VITAL subset.
Importantly, statistical significance reflects detectability of differences rather than their magnitude.
}
\centering
\small
\renewcommand{\arraystretch}{1.2}
\begin{tabular}{lcccc}
\toprule
\textbf{Method} & \textbf{Dataset} & \textbf{F-statistic} & \textbf{p-value} & \textbf{Significance} \\
\midrule
Direct Prompting      & GlobalOpinionQA & 37.94 & $< 10^{-6}$ & Significant \\
Direct Prompting      & VITAL           & 4.25  & 0.014       & Significant \\
\midrule
Modular Pluralism     & GlobalOpinionQA & 38.02 & $< 10^{-6}$ & Significant \\
Modular Pluralism     & VITAL           & 2.78  & 0.062       & Not Significant \\
\midrule
\textsc{Demographic Pluralism} & GlobalOpinionQA & 19.15 & $< 10^{-6}$ & Significant \\
\textsc{Demographic Pluralism} & VITAL           & 31.56 & $< 10^{-6}$ & Significant \\
\bottomrule
\end{tabular}

\label{tab:anova-resource}
\end{table*}

This section provides detailed statistics on alignment performance across country resource levels, supplementing the analysis in Section~\ref{sec:resource_impact}.
Table~\ref{tab:resource_level_jsd} reports average Jensen--Shannon Distance (JSD) across high-, medium-, and low-resource country groups, while Table~\ref{tab:anova-resource} presents one-way ANOVA results testing the statistical significance of these differences.

Across both benchmarks, \textsc{Demographic Pluralism} achieves the lowest JSD within every resource tier.
On GlobalOpinionQA, our method reduces JSD by 8.8\%--15.6\% compared to Modular Pluralism across all tiers (e.g., 0.2982 vs.\ 0.3268 in the low-resource group).
On VITAL, improvements range from 3.4\% in low-resource countries to 16.1\% in high-resource countries.
Notably, \textsc{Demographic Pluralism} maintains its advantage even in the low-resource tier, where alignment is generally most challenging.

The ANOVA results reveal that socioeconomic context systematically influences alignment difficulty.
On GlobalOpinionQA, all three methods exhibit statistically significant variation across resource tiers ($F > 19$, $p < 10^{-6}$).
On VITAL, both Direct Prompting ($F = 4.25$, $p = 0.014$) and \textsc{Demographic Pluralism} ($F = 31.56$, $p < 10^{-6}$) show significant differences, whereas Modular Pluralism does not ($F = 2.78$, $p = 0.062$).
While no method fully eliminates resource-level disparities, \textsc{Demographic Pluralism} achieves lower JSD than both baselines in every resource tier.

\section{Robustness to Intermediate Model Scale}
\label{sec:appendix-scaling}

We evaluate whether \textsc{Demographic Pluralism}'s gains depend on the capacity of the intermediate generator that produces demographic profiles and perspectives. We hold the downstream opinion-distribution model fixed at Qwen2.5-7B-Instruct and swap the intermediate generator from Mistral-7B-Instruct to Qwen3-32B, a larger and more recent model.

\begin{table*}[t]
\caption{
Alignment (JSD $\downarrow$) with Qwen2.5-7B-Instruct as the downstream opinion model. Replacing the Mistral-7B intermediate generator with Qwen3-32B does not improve \textsc{Demographic Pluralism}; each configuration remains substantially better than CoT using the same intermediate generator.
}
\centering
\small
\renewcommand{\arraystretch}{1.25}
\setlength{\tabcolsep}{6pt}
\begin{tabular}{llcc}
\toprule
\textbf{Method} & \textbf{Intermediate generator}
& \textbf{Global JSD $\downarrow$}
& \textbf{VITAL JSD $\downarrow$} \\
\midrule
CoT & Mistral-7B & 0.3893 & 0.3846 \\
CoT & Qwen3-32B & 0.4788 & 0.4504 \\
\midrule
\textsc{Demographic Pluralism} & Mistral-7B & \textbf{0.2681} & \textbf{0.2547} \\
\textsc{Demographic Pluralism} & Qwen3-32B & 0.2788 & 0.3095 \\
\bottomrule
\end{tabular}
\label{tab:qwen3-scaling}
\end{table*}

Table~\ref{tab:qwen3-scaling} reports JSD on both benchmarks. Both \textsc{Demographic Pluralism} configurations outperform CoT using the same intermediate generator. However, replacing Mistral-7B with Qwen3-32B does not improve \textsc{Demographic Pluralism}: JSD rises by 0.0107 on GlobalOpinionQA and 0.0548 on VITAL. The artifact analysis below tests whether this absence of improvement can be explained by similar intermediate outputs.

\begin{table*}[t]
\caption{
Semantic similarity of generated artifacts between Mistral-7B and Qwen3-32B (sentence embeddings, cosine similarity), across 665{,}264 matched (question, demographic group) pairs.
Cross-model similarity is moderate ($\sim$0.50), and Qwen3-32B produces more numerous and internally diverse perspectives, yet does not improve downstream alignment.
}
\centering
\small
\renewcommand{\arraystretch}{1.2}
\setlength{\tabcolsep}{5pt}
\begin{tabular}{lcc}
\toprule
\textbf{Metric} & \textbf{Global} & \textbf{VITAL} \\
\midrule
Cross-model perspective sim.    & $0.500 \pm 0.115$ & $0.540 \pm 0.111$ \\
Cross-model background sim.     & $0.594 \pm 0.144$ & $0.599 \pm 0.137$ \\
Within-Mistral perspective sim. & $0.767 \pm 0.116$ & $0.788 \pm 0.109$ \\
Within-Qwen perspective sim.    & $0.575 \pm 0.121$ & $0.587 \pm 0.123$ \\
\midrule
Mistral avg.\ perspectives/group & 2.73 & 2.81 \\
Qwen3-32B avg.\ perspectives/group & 2.99 & 2.99 \\
\bottomrule
\end{tabular}
\label{tab:artifact-similarity}
\end{table*}

To characterize \emph{how} the two intermediate models differ, we compute cosine similarity with all-MiniLM-L6-v2 sentence embeddings~\citep{reimers2019sentence}\footnote{\url{https://huggingface.co/sentence-transformers/all-MiniLM-L6-v2}} across 665{,}264 matched (question, demographic group) pairs. Table~\ref{tab:artifact-similarity} shows that cross-model similarity is moderate ($\sim$0.50 for perspectives, $\sim$0.60 for background descriptions)---substantively different artifacts, not paraphrases. Within-Mistral perspective similarity (0.77--0.79) is much higher than within-Qwen (0.57--0.59), and Qwen3-32B generates slightly more perspectives per demographic group (2.99 vs.\ 2.73--2.81). Thus Qwen3-32B produces more numerous and more internally diverse perspectives, but these richer artifacts do not improve downstream alignment. This result indicates that increased generator scale alone is not responsible for the framework's gains.


\subsection{Qualitative Comparison of Intermediate Artifacts}
\label{sec:appendix-artifact-comparison}

The quantitative scaling analysis in Appendix~\ref{sec:appendix-scaling} shows that replacing Mistral-7B-Instruct with Qwen3-32B does not improve downstream alignment: JSD increases from 0.2681 to 0.2788 on GlobalOpinionQA and from 0.2547 to 0.3095 on VITAL.
This result raises a natural question: why do richer artifacts from the larger generator fail to improve the final estimates, and what role does the structural demographic scaffolding play?
To answer this, we conduct a detailed qualitative comparison of intermediate artifacts produced by both models for matched demographic groups, complementing the cosine-similarity statistics in Table~\ref{tab:artifact-similarity}.

The two models produce substantively different perspectives---different names, backgrounds, reasoning styles, and emphasis---yet both cover a similar opinion space for each demographic group.
Cross-model perspective cosine similarity centers around 0.50 (GlobalOpinionQA) and 0.54 (VITAL), confirming that the artifacts are not paraphrases but genuinely distinct articulations.
Despite this divergence, Qwen3-32B does not improve downstream JSD (Table~\ref{tab:qwen3-scaling}), showing that richer intermediate text alone is insufficient to improve the final estimates.

\paragraph{Quantitative artifact-level statistics.}
Table~\ref{tab:artifact-similarity} reports cross-model and within-model similarity metrics computed over all 665{,}264 matched question--group pairs using all-MiniLM-L6-v2 sentence embeddings.

Three patterns emerge: (i)~cross-model perspective similarity (${\sim}0.50$) is substantially lower than within-model similarity for either model, indicating genuinely different articulations rather than surface paraphrases; (ii)~within-Mistral similarity (0.77--0.79) is much higher than within-Qwen (0.57--0.59), meaning Qwen3-32B generates more internally diverse perspectives per demographic group; and (iii)~background similarity (${\sim}0.60$) exceeds perspective similarity (${\sim}0.50$), confirming that both models describe the same demographic profile comparably but diverge more in opinion articulation.
Despite these substantial differences in intermediate artifacts, the larger generator does not improve downstream alignment (Table~\ref{tab:qwen3-scaling}), reinforcing that artifact richness alone does not explain the framework's gains.

\paragraph{Worked Example 1: Belgium / GlobalOpinionQA (EU influence question).}

We present a full side-by-side comparison for a representative GlobalOpinionQA question asked in Belgium.
The demographic group is: \textit{15--64 years old male, tertiary education, Christian}.

\begin{quote}
\small
\textbf{Question:} ``When it comes to Germany's decision-making in the European Union, do you think Germany has too much influence, has too little influence or has about the right amount of influence?'' \\
\textbf{Options:} A.~Has too much influence \quad B.~Has too little influence \quad C.~Has about the right amount of influence \quad D.~DK/Refused
\end{quote}

\noindent\textbf{Mistral-7B-Instruct-v0.1:}

\begin{quote}
\small
\textit{Name:} Pieter

\textit{Background:} Pieter is a 35-year-old Belgian male with a tertiary education. He is a devout Christian and attends church regularly. Pieter works as a software engineer in Brussels and enjoys spending time with his family. He is politically moderate and values social justice and equality.

\textit{Perspective 1:} Pieter believes that Germany's influence in the EU is too great, as it often dominates decision-making and fails to consider the needs of smaller member states like Belgium.

\textit{Perspective 2:} Pieter sees Germany's influence in the EU as just right, as it is a major player but also recognizes the importance of collaboration with other member states. He believes that Germany has a responsibility to use its influence for the greater good of the EU.
\end{quote}

\noindent\textbf{Qwen3-32B:}

\begin{quote}
\small
\textit{Name:} Jean Van den Berghe

\textit{Background:} Jean is a 34-year-old civil engineer working for a mid-sized infrastructure firm in Brussels. He grew up in a small town in Flanders and attended a Catholic secondary school before pursuing a master's degree in engineering at KU Leuven. He is married with two children and actively involved in his local parish, where he volunteers with youth outreach programs. Jean identifies as a practicing Catholic and values the role of religion in personal ethics and community life. His tertiary education and professional career are typical of a middle-class Belgian male in his demographic group, and his lifestyle reflects a balance between professional ambition and religious commitment.

\textit{Perspective 1:} Germany has about the right amount of influence in the EU because it's the largest economy and should naturally play a leading role in shaping decisions that affect all member states.

\textit{Perspective 2:} I think Germany has too much influence in the EU, especially since Belgium often ends up implementing policies we had little say in, which can feel frustrating for smaller countries.

\textit{Perspective 3:} Germany's influence in the EU is necessary but should be balanced with the voices of smaller nations like Belgium to ensure fair decision-making for everyone.
\end{quote}

\paragraph{Observations (Example 1).}
Both models cover the same opinion space---``too much influence'' and ``about the right amount''---but with markedly different surface articulations.
Qwen3-32B adds a third nuanced perspective and provides a substantially richer background with specific cultural details (KU Leuven, Flanders, parish involvement), while Mistral-7B produces a more generic profile.
The perspectives use different framing: Mistral speaks in third person about ``Pieter,'' while Qwen uses first person.
Despite these differences, both capture the range of plausible opinions for this demographic group, and both would contribute similar probability mass to the aggregated distribution.

\paragraph{Worked Example 2: Argentina / VITAL (air pollution question).}

We present a second full comparison for a VITAL question asked in Argentina.
The demographic group is: \textit{15--64 years old male, tertiary education, White/Castizo}.

\begin{quote}
\small
\textbf{Question:} ``Now I am going to read you a list of things that may be problems in our country. As I read each one, please tell me if you think it is a very big problem, a moderately big problem, a small problem or not a problem at all\ldots\ Air pollution'' \\
\textbf{Options:} A.~Very big problem \quad B.~Moderately big problem \quad C.~Small problem \quad D.~Not a problem at all \quad E.~DK/Refused
\end{quote}

\noindent\textbf{Mistral-7B-Instruct-v0.1:}

\begin{quote}
\small
\textit{Name:} Juan

\textit{Background:} Juan is a 35-year-old man from Buenos Aires, Argentina. He has a Bachelor's degree in Business Administration and works as a marketing manager at a multinational corporation. Juan is of White/Castizo descent and speaks Spanish fluently. He enjoys spending time with his family, playing soccer, and traveling to different countries. Juan is an active member of his community and volunteers at a local charity organization.

\textit{Perspective 1:} Air pollution is a very big problem in Argentina, especially in Buenos Aires where I live. It affects our health and the environment. We need to take immediate action to reduce emissions and improve air quality.

\textit{Perspective 2:} I don't think air pollution is a big problem in Argentina. It's not something that directly affects me or my community. I believe other issues, like poverty and crime, should be our main concerns.

\textit{Perspective 3:} Air pollution is a moderately big problem in Argentina. It affects our health and the environment, but it's not the most pressing issue we face. We should focus on improving our economy and creating more job opportunities.
\end{quote}

\noindent\textbf{Qwen3-32B:}

\begin{quote}
\small
\textit{Name:} Javier Morales

\textit{Background:} Javier is a middle-class professional living in Buenos Aires. He studied economics at the University of Buenos Aires and currently works as a financial analyst for a multinational corporation. He was born and raised in a traditional White/Castizo family in the suburbs of the capital. His parents were both public school teachers, and he grew up in a stable, nuclear family environment. Javier is fluent in Spanish and English, and he is well integrated into the urban professional culture of Argentina. He is married with one child and is active in local community and professional networks.

\textit{Perspective 1:} Air pollution is a moderately big problem in Buenos Aires, especially in industrial areas, but it doesn't affect my daily life significantly due to where I live and work.

\textit{Perspective 2:} I don't see air pollution as a major issue because the government has implemented effective environmental policies in recent years to control emissions and promote green initiatives.

\textit{Perspective 3:} While I acknowledge that air pollution exists, I believe it's a small problem compared to other national challenges like economic instability and unemployment.
\end{quote}

\paragraph{Observations (Example 2).}
Both models generate three perspectives spanning the severity spectrum, but with different emphasis.
Mistral-7B includes a ``very big problem'' perspective while Qwen3-32B skews moderate---none of its perspectives rate air pollution as ``very big.''
The reasoning differs substantially: Mistral cites health and environmental impacts directly, while Qwen contextualizes the issue relative to economic concerns and government policy.
Both backgrounds describe a similar professional profile in Buenos Aires but with different specifics (marketing manager vs.\ financial analyst; Business Administration vs.\ economics at UBA).
Despite these divergent framings, both sets of perspectives cover overlapping answer tendencies; the richer Qwen3-32B artifacts nevertheless do not improve aggregate JSD.

\paragraph{Interpretation: scaffolding over artifact quality.}
The worked examples, together with the quantitative statistics in Table~\ref{tab:artifact-similarity}, support three conclusions:

\begin{enumerate}
\item \textbf{Qwen3-32B produces more diverse and nuanced perspectives.} Within-model similarity is substantially lower for Qwen (0.575--0.587) than for Mistral (0.767--0.788), and Qwen generates slightly more perspectives per group (2.99 vs.\ 2.73--2.81). The qualitative examples confirm this: Qwen provides richer backgrounds with specific cultural details and adds additional nuanced viewpoints.

\item \textbf{Despite this diversity gap, both models cover convergent opinion spaces.} For each demographic group, both models generate perspectives spanning the plausible range of responses. Cross-model similarity of ${\sim}0.50$ reflects different \emph{articulations} of similar \emph{positions}, not fundamentally different opinion coverage.

\item \textbf{Richer artifacts alone do not improve downstream alignment.} The framework conditions perspective generation on fine-grained demographically grounded groups and averages distributions within groups before applying the evaluated cross-group weighting rule. In this experiment, Mistral-7B produces more homogeneous perspectives yet lower JSD than Qwen3-32B, indicating that increased textual sophistication is not sufficient for better population-level estimates.
\end{enumerate}

This finding has practical implications: the smaller Mistral generator is more cost-efficient and performs at least as well in this comparison. It also reinforces the role of structured demographic grounding and aggregation while showing that unconstrained increases in perspective detail do not guarantee better population-level opinion distributions.

\paragraph{Relation to Appendix~\ref{sec:appendix-scaling}.}
This qualitative analysis directly extends the quantitative scaling experiment in Appendix~\ref{sec:appendix-scaling}, which established that Qwen3-32B artifacts do not improve downstream JSD over Mistral-7B artifacts.
The present subsection provides the mechanistic explanation: while the artifacts differ substantively (cosine ${\sim}0.50$), demographic scaffolding ensures convergent opinion-space coverage regardless of intermediate-model capacity.
The main body discusses this scaling result in Section~\ref{sec:scaling}; this subsection provides qualitative context for the measured difference.

\section{Runtime Analysis}
\label{appendix:runtime}

\begin{table*}[t]
\centering
\small
\renewcommand{\arraystretch}{1.2}
\caption{Wall-clock time (seconds) decomposed into artifact generation and opinion generation. All methods use Mistral-7B-Instruct on a single A100 GPU.}
\label{tab:runtime}
\setlength{\tabcolsep}{6pt}
\begin{tabular}{llrrr}
\toprule
\textbf{Benchmark} & \textbf{Method} & \textbf{Artifact (s)} & \textbf{Opinion (s)} & \textbf{Total (s)} \\
\midrule
\multirow{4}{*}{\shortstack[l]{GlobalOpinionQA\\($n$=28{,}763)}}
& Direct Prompting & 0 & 854 & 854 \\
& CoT & 73{,}101 & 461 & 73{,}562 \\
& Modular Pluralism & 48{,}402 & 11{,}454 & 59{,}856 \\
& \textsc{Demographic Pluralism} & 45{,}578 & 16{,}306 & 61{,}884 \\
\midrule
\multirow{4}{*}{\shortstack[l]{VITAL\\($n$=1{,}676)}}
& Direct Prompting & 0 & 92 & 92 \\
& CoT & 4{,}771 & 71 & 4{,}842 \\
& Modular Pluralism & 1{,}475 & 711 & 2{,}185 \\
& \textsc{Demographic Pluralism} & 5{,}510 & 1{,}324 & 6{,}834 \\
\bottomrule
\end{tabular}
\end{table*}

Table~\ref{tab:runtime} reports measured wall-clock time for the full pipeline. On GlobalOpinionQA, \textsc{Demographic Pluralism} takes 61{,}884 seconds, comparable to Modular Pluralism (59{,}856) and below CoT (73{,}562); artifact generation accounts for 73.6\% of its cost.

On the smaller VITAL subset, its 6{,}834-second runtime exceeds CoT (4{,}842) and Modular Pluralism (2{,}185), largely because it queries each perspective separately. Those completed distributions can be reweighted without further inference. In addition, because population weights are known after group construction, direct top-$K$ generation can skip unselected perspective and opinion calls. Figure~\ref{fig:minimal_perspectives} establishes the quality retained by this selection, but its end-to-end runtime has not been measured.

\section{Demographic Weight and Group-Level Prediction Error}
\label{appendix:aggregation-bias}

Equal- or inverse-weighted aggregation matches or outperforms weighted aggregation in every main-table setting, with equal-weighted aggregation strongest overall (Table~\ref{tab:alignment-results}). We test whether demographic weight is associated with group-level prediction error and whether that relationship varies by country resource tier. Using Qwen2.5-7B-Instruct, we compute JSD between each group's mean prediction and the country-level reference distribution, then correlate JSD with normalized demographic weight. This yields 917{,}538 observations on GlobalOpinionQA and 55{,}298 on VITAL. The analysis identifies an aggregation-relevant association; it is not designed to determine the cause of group-level error.

\paragraph{Results.}
Table~\ref{tab:weight-bias-correlation} reports Spearman rank correlations between per-group demographic weight and per-group prediction error, stratified by country resource level. On GlobalOpinionQA, the correlation is strongest in low-resource countries ($\rho = 0.059$, $p = 1.3 \times 10^{-95}$), where the highest-weight third of demographic groups has 13.8\% higher JSD than the lowest-weight third (0.1295 vs.\ 0.1138). In high-resource countries, the gap is only 1.8\%. On VITAL, the association is positive across all tiers and largest for medium-resource countries ($\rho = 0.137$), where the JSD gap between terciles is 20.5\%. The association is not uniform across tiers, but its presence in low-resource GlobalOpinionQA countries and all VITAL tiers shows that it is not confined to one benchmark.

\begin{table*}[t]
\caption{
Spearman correlation ($\rho$) between per-group demographic weight and per-group prediction error (JSD) on GlobalOpinionQA and VITAL (Qwen2.5-7B-Instruct), stratified by country resource level. Positive $\rho$ indicates that high-weight demographic groups produce less accurate predictions. The rightmost columns report mean JSD for the lowest- and highest-weight terciles.
}
\centering
\small
\renewcommand{\arraystretch}{1.2}
\begin{tabular}{llccccc}
\toprule
\textbf{Benchmark} & \textbf{Resource Tier} & \textbf{N (groups)} & \textbf{Spearman $\rho$} & \textbf{$p$-value} & \textbf{JSD (bot.\ 1/3)} & \textbf{JSD (top 1/3)} \\
\midrule
\multirow{3}{*}{GlobalOpinionQA}
& High    & 436{,}145 & $-0.006$ & $1.7 \times 10^{-4}$ & 0.1300 & 0.1323 \\
& Medium  & 359{,}202 & $+0.010$ & $1.0 \times 10^{-9}$ & 0.1329 & 0.1369 \\
& Low     & 122{,}191 & $+0.059$ & $1.3 \times 10^{-95}$ & 0.1138 & 0.1295 \\
\midrule
\multirow{3}{*}{VITAL}
& High    & 22{,}449 & $+0.065$ & $1.4 \times 10^{-22}$ & 0.0965 & 0.1131 \\
& Medium  & 23{,}698 & $+0.137$ & $1.0 \times 10^{-99}$ & 0.0922 & 0.1111 \\
& Low     & 9{,}151 & $+0.085$ & $4.4 \times 10^{-16}$ & 0.1063 & 0.1216 \\
\bottomrule
\end{tabular}
\label{tab:weight-bias-correlation}
\end{table*}

\paragraph{Interpretation.}
The measured association offers one explanation for the aggregation result: weighted aggregation gives more influence to groups with larger observed errors, whereas equal-weighted aggregation limits that concentration. The association is strongest in low-resource GlobalOpinionQA countries and appears across all VITAL tiers. These data do not identify why high-weight groups are harder to predict; pretraining coverage, demographic heterogeneity, and reference-distribution noise remain untested explanations. Equal-weighted aggregation performs best overall in these experiments, but should not be interpreted as a universally correct population estimator.


\section{Metric Comparison: JSD, EMD, and Hybrid}
\label{sec:appendix-metric-comparison}

The main paper reports Jensen--Shannon distance because it applies uniformly to nominal and ordinal response scales.
However, 59.3\% of GlobalOpinionQA questions and 55.4\% of VITAL questions have ordinal response options, for which the distance between answer choices carries useful information.
We therefore additionally report normalized Earth Mover's Distance over all evaluated instances as a sensitivity analysis.
For ordinal questions, EMD distinguishes probability mass moved between nearby options from mass moved across the response scale; for nominal questions, the option order has no inherent geometric meaning.
To retain the appropriate structure for each question, we also report a question-aware Hybrid metric that uses EMD for ordinal questions and JSD for nominal questions.

\paragraph{Metric definitions.}
Let $P$ and $\hat{P}$ denote normalized human and predicted distributions over $K$ response options, and let $M=(P+\hat{P})/2$.
We use JSD to denote the Jensen--Shannon \emph{distance}, computed with natural logarithms:
\begin{equation}
\begin{aligned}
    d_{\mathrm{JS}}^2(P,\hat{P})
    &= \tfrac{1}{2}D_{\mathrm{KL}}(P\Vert M) \\
    &\quad + \tfrac{1}{2}D_{\mathrm{KL}}(\hat{P}\Vert M).
\end{aligned}
\end{equation}
JSD ignores option order, allowing the same calculation to be applied to both nominal and ordinal questions.

For EMD, we remove non-substantive options such as ``Don't know'' and ``Refused,'' renormalize both distributions, and place the remaining $K$ options at equally spaced positions $x_i=(i-1)/(K-1)$ on $[0,1]$ while preserving their dataset order.
The resulting one-dimensional normalized Earth Mover's Distance is
\begin{equation}
    d_{\mathrm{EMD}}(P,\hat{P}) =
    \frac{1}{K-1}
    \sum_{i=1}^{K-1}
    \left|\sum_{j=1}^{i}(P_j-\hat{P}_j)\right|.
\end{equation}
The full EMD results average this value over all evaluated question--country instances.
For ordinal questions, the option positions represent a semantic response scale, so EMD penalizes confusion between distant choices more heavily than confusion between nearby choices.
For nominal questions, this geometry is arbitrary, which motivates the question-aware Hybrid metric.

\paragraph{Question-aware Hybrid metric.}
Let $\mathcal{O}$ and $\mathcal{N}$ denote the sets of ordinal and nominal questions, and let $\mathcal{I}$ be the set of evaluated question--country instances.
For each instance $n\in\mathcal{I}$, with question $q_n$, we compute
\begin{equation}
    d_{\mathrm{Hybrid}}(n) =
    \begin{cases}
        d_{\mathrm{EMD}}(P_n,\hat{P}_n), & q_n \in \mathcal{O}, \\
        d_{\mathrm{JS}}(P_n,\hat{P}_n), & q_n \in \mathcal{N}.
    \end{cases}
\end{equation}
The reported Hybrid score is the macro-average $|\mathcal{I}|^{-1}\sum_{n\in\mathcal{I}}d_{\mathrm{Hybrid}}(n)$ over evaluated question--country instances.
It is not the arithmetic mean of aggregate JSD and EMD.

\paragraph{Ordinal versus nominal classification.}
We classify a question as ordinal when its substantive response options form a graded scale, such as agreement, intensity, quality, or frequency.
Deterministic matching over option text identifies explicit degree modifiers (e.g., ``strongly,'' ``mostly,'' ``very,'' and ``somewhat'') and frequency terms (e.g., ``always,'' ``often,'' ``sometimes,'' and ``never''); the metric configuration records the substantive options in their semantic order.
Questions whose options have no inherent order, including named entities and policy alternatives, use JSD.

\paragraph{Full-corpus comparison.}
Table~\ref{tab:metric-triple} reports all three metrics for the full-corpus protocol across four backbones and two evaluation sets.
We include both WEIGHTED and EQUAL-WEIGHTED aggregation to show that the comparison with the baselines that do not use observed opinion distributions as training data does not depend on a single aggregation choice.

\begin{table*}[t]
\caption{
Alignment under JSD, EMD, and the question-aware Hybrid metric (all $\downarrow$) in the full-corpus protocol. \textbf{Bold} marks the best result, and \underline{underlining} marks the second-best result within each backbone, evaluation set, and metric.
}
\label{tab:metric-triple}
\centering
\large
\renewcommand{\arraystretch}{2.0}
\setlength{\tabcolsep}{2.8pt}
\resizebox{\textwidth}{!}{%
\begin{tabular}{l ccc ccc ccc ccc ccc ccc ccc ccc}
\toprule
& \multicolumn{6}{c}{\textbf{Mistral-7B-Instruct}} & \multicolumn{6}{c}{\textbf{Phi-3.5-Instruct}} & \multicolumn{6}{c}{\textbf{Qwen2.5-7B-Instruct}} & \multicolumn{6}{c}{\textbf{Qwen3-14B}} \\
\cmidrule(lr){2-7} \cmidrule(lr){8-13} \cmidrule(lr){14-19} \cmidrule(lr){20-25}
& \multicolumn{3}{c}{Global} & \multicolumn{3}{c}{VITAL} & \multicolumn{3}{c}{Global} & \multicolumn{3}{c}{VITAL} & \multicolumn{3}{c}{Global} & \multicolumn{3}{c}{VITAL} & \multicolumn{3}{c}{Global} & \multicolumn{3}{c}{VITAL} \\
\cmidrule(lr){2-4} \cmidrule(lr){5-7} \cmidrule(lr){8-10} \cmidrule(lr){11-13} \cmidrule(lr){14-16} \cmidrule(lr){17-19} \cmidrule(lr){20-22} \cmidrule(lr){23-25}
\textbf{Method} & JSD & EMD & Hyb & JSD & EMD & Hyb & JSD & EMD & Hyb & JSD & EMD & Hyb & JSD & EMD & Hyb & JSD & EMD & Hyb & JSD & EMD & Hyb & JSD & EMD & Hyb \\
\midrule
Direct Prompting
& .3224 & .2535 & .2661
& .3645 & .2847 & .2988
& .4178 & .2967 & .3116
& .3925 & .2768 & .3116
& .3906 & .2871 & .3004
& .3576 & .2380 & .2732
& .4076 & .3033 & .3153
& .3412 & .2312 & .2521 \\
CoT
& .3905 & .2905 & .3118
& .3862 & .2754 & .3039
& .4806 & .3307 & .3558
& .4639 & .3131 & .3426
& .3893 & .2857 & .2980
& .3846 & .2670 & .2876
& .4234 & .3008 & .3160
& .4025 & .2761 & .3006 \\
Modular Pluralism
& .3292 & .2489 & .2670
& .3574 & .2857 & .3065
& .3366 & .2341 & .2512
& .3226 & .2445 & .2537
& .3551 & .2371 & .2598
& .3462 & .2180 & .2636
& .3369 & .2328 & .2522
& .3044 & .2106 & .2368 \\
\midrule
\textsc{Ours} (WEIGHTED)
& \underline{.2884} & \underline{.2372} & \underline{.2437}
& \textbf{.3275} & \textbf{.2750} & \textbf{.2835}
& \underline{.2749} & \underline{.2061} & \underline{.2127}
& \underline{.2843} & \textbf{.2178} & \textbf{.2285}
& \underline{.2750} & \underline{.2113} & \underline{.2106}
& \underline{.2588} & \underline{.1930} & \underline{.2062}
& \underline{.2835} & \underline{.2130} & \underline{.2131}
& \underline{.2461} & \underline{.1867} & \underline{.1981} \\
\textsc{Ours} (EQUAL-WT)
& \textbf{.2869} & \textbf{.2356} & \textbf{.2429}
& \underline{.3276} & \underline{.2753} & \underline{.2840}
& \textbf{.2669} & \textbf{.2026} & \textbf{.2092}
& \textbf{.2808} & \underline{.2222} & \underline{.2303}
& \textbf{.2681} & \textbf{.2018} & \textbf{.2072}
& \textbf{.2547} & \textbf{.1910} & \textbf{.2051}
& \textbf{.2757} & \textbf{.2085} & \textbf{.2086}
& \textbf{.2386} & \textbf{.1802} & \textbf{.1928} \\
\bottomrule
\end{tabular}%
}
\end{table*}

\paragraph{Question-disjoint EMD and Hybrid comparison.}
Tables~\ref{tab:heldout-emd} and~\ref{tab:heldout-hybrid} mirror the two question-disjoint scopes in Table~\ref{tab:heldout-response-supervised} under normalized EMD and the question-aware Hybrid metric, respectively.
As in the main table, we report only EQUAL-WEIGHTED \textsc{Demographic Pluralism}.

\begin{table*}[t]
\caption{
Normalized EMD ($\downarrow$) on question-disjoint GlobalOpinionQA test sets. Panel A covers all 131 test countries; Panel B uses the 19-country P2P-compatible subset. \textbf{Bold} marks the best result within each backbone and panel.
}
\label{tab:heldout-emd}
\centering
\small
\renewcommand{\arraystretch}{1.08}
\setlength{\tabcolsep}{7pt}
\begin{tabular}{lcccc}
\toprule
\textbf{Method} & \textbf{Mistral-7B} & \textbf{Phi-3.5} & \textbf{Qwen2.5-7B} & \textbf{Qwen3-14B} \\
\midrule
\multicolumn{5}{l}{\textit{Panel A: 131-country question-disjoint test (15{,}893 instances, 1{,}191 questions)}} \\
Direct Prompting & .256 & .271 & .252 & .257 \\
CoT & .295 & .314 & .249 & .262 \\
Modular Pluralism & .213 & .203 & .194 & .196 \\
Cao et al. & \textbf{.149} & .267 & .361 & .238 \\
\textsc{Demographic Pluralism} & .211 & \textbf{.181} & \textbf{.169} & \textbf{.169} \\
\midrule
\multicolumn{5}{l}{\textit{Panel B: P2P-compatible test (3{,}669 instances, 860 questions, 19 countries)}} \\
Direct Prompting & .251 & .258 & .245 & .255 \\
CoT & .297 & .303 & .248 & .255 \\
Modular Pluralism & .208 & .197 & .187 & .190 \\
Cao et al. & \textbf{.144} & .271 & .356 & .236 \\
Prompts to Proxies & .223 & .178 & .179 & .169 \\
\textsc{Demographic Pluralism} & .203 & \textbf{.174} & \textbf{.161} & \textbf{.164} \\
\bottomrule
\end{tabular}
\end{table*}

\begin{table*}[t]
\caption{
Question-aware Hybrid distance ($\downarrow$) on the same question-disjoint test sets. The metric uses EMD for ordinal questions and JSD for nominal questions. \textbf{Bold} marks the best result within each backbone and panel.
}
\label{tab:heldout-hybrid}
\centering
\small
\renewcommand{\arraystretch}{1.08}
\setlength{\tabcolsep}{7pt}
\begin{tabular}{lcccc}
\toprule
\textbf{Method} & \textbf{Mistral-7B} & \textbf{Phi-3.5} & \textbf{Qwen2.5-7B} & \textbf{Qwen3-14B} \\
\midrule
\multicolumn{5}{l}{\textit{Panel A: 131-country question-disjoint test (15{,}893 instances, 1{,}191 questions)}} \\
Direct Prompting & .285 & .323 & .313 & .317 \\
CoT & .334 & .378 & .307 & .329 \\
Modular Pluralism & .253 & .247 & .244 & .247 \\
Cao et al. & \textbf{.180} & .305 & .437 & .271 \\
\textsc{Demographic Pluralism} & .247 & \textbf{.218} & \textbf{.210} & \textbf{.215} \\
\midrule
\multicolumn{5}{l}{\textit{Panel B: P2P-compatible test (3{,}669 instances, 860 questions, 19 countries)}} \\
Direct Prompting & .283 & .314 & .310 & .323 \\
CoT & .339 & .371 & .310 & .328 \\
Modular Pluralism & .251 & .244 & .239 & .245 \\
Cao et al. & \textbf{.177} & .313 & .438 & .272 \\
Prompts to Proxies & .349 & .313 & .291 & .278 \\
\textsc{Demographic Pluralism} & .242 & \textbf{.213} & \textbf{.204} & \textbf{.213} \\
\bottomrule
\end{tabular}
\end{table*}

Under both EMD and Hybrid, \textsc{Demographic Pluralism} outperforms Cao et al. on three of four backbones in both held-out scopes and outperforms P2P on all four backbones in the compatible subset.
Cao et al. remains best on Mistral-7B.

\paragraph{Cross-metric agreement.}
Across the full-corpus results, \textsc{Demographic Pluralism} is best or second-best for every backbone, evaluation set, and metric.
The baseline ordering is also stable in most configurations, with Modular Pluralism generally outperforming Direct Prompting and CoT.
The remaining rank changes occur primarily between WEIGHTED and EQUAL-WEIGHTED aggregation, whose scores differ by less than 0.010 within each cell.

\paragraph{Interpreting metric values.}
JSD and EMD encode different geometries and have different numerical scales, so their absolute magnitudes should not be compared directly.
The relevant evidence is the agreement in method rankings and conclusions within each metric.
EMD additionally distinguishes nearby from distant errors on ordinal scales, while Hybrid avoids imposing that structure on nominal questions.

\paragraph{Conclusion.}
The EMD and question-aware Hybrid results preserve the main JSD conclusion: modeling demographic structure and within-group variation improves alignment over baselines that do not use opinion-distribution training data across backbones and evaluation sets.
Hybrid is the most appropriate secondary metric for benchmarks containing both nominal and ordinal questions because it respects each question's response structure.

\end{document}